\documentclass[lettersize,journal]{IEEEtran}

\ifCLASSOPTIONcompsoc
\usepackage[nocompress]{cite}
\else
\usepackage{cite}
\fi

\ifCLASSINFOpdf
\else
\fi

\usepackage[table]{xcolor}
\definecolor{Gray}{gray}{0.9}
\usepackage{gensymb}

\usepackage{amsmath,amssymb,amsfonts}
\usepackage{graphicx}
\usepackage{textcomp}
\usepackage{tikz}
\usetikzlibrary{bayesnet}

\usepackage{mathrsfs}
\usepackage[colorlinks,linkcolor=red,citecolor=blue]{hyperref}
\usepackage{amsmath}

\usepackage{amsfonts,subfigure}
\usepackage{amssymb}
\usepackage{multirow}
\usepackage{bm}
\usepackage{sansmath}
\usepackage{booktabs}
\usepackage[ruled]{algorithm2e}

\usepackage{graphicx}
\usepackage{amsthm}

\graphicspath{{figure/}}

\usepackage{bm}
\usepackage{color,soul}

\usepackage{tikz}

\usepackage{multirow}
\usepackage{adjustbox}
\usepackage{environ}
\usepackage{setspace}

\usepackage{enumitem}

\newcommand{\tableref}[1]{Table~\ref{#1}}
  
\newcommand{\figref}[1]{Figure~\ref{#1}} 
\graphicspath{{image/}}

\usepackage{makecell}
\definecolor{Gray}{gray}{0.9}
\usepackage{graphicx}

\begin{document}
	
	\title{
		Unveiling the Inflexibility of Adaptive Embedding in Traffic Forecasting }
	
	%
	%
	%
	%
	

	\author{Hongjun~Wang,
	Jiyuan~Chen,
	Lingyu~Zhang,
	Renhe~Jiang,
	and
	Xuan Song
	\IEEEcompsocitemizethanks{
		\IEEEcompsocthanksitem Hongjun Wang, Jiyuan Chen and Lingyu Zhang are with (1) SUSTech-UTokyo Joint Research Center on Super Smart City, Department of Computer Science and Engineering
		(2) Research Institute of Trustworthy Autonomous Systems, Southern University of Science and Technology (SUSTech), Shenzhen, China.
		E-mail: {wanghj2020,11811810}@mail.sustech.edu.cn, and zhanglingyu@didiglobal.com. 
		\IEEEcompsocthanksitem Xuan Song is with (1) School of Artificial Intelligence, Jilin University  (2) Research Institute of Trustworthy Autonomous Systems, Southern University of Science and Technology (SUSTech), Shenzhen, China. Email: songxuan@jlu.edu.cn.
		\IEEEcompsocthanksitem R. Jiang is with Center for Spatial Information Science, University of
		Tokyo, Tokyo, Japan. Email: jiangrh@csis.u-tokyo.ac.jp.
		\IEEEcompsocthanksitem Corresponding to  Xuan Song;
	}
}

	%
	%

\markboth{Journal of \LaTeX\ Class Files,~Vol.~XX, No.~X, August~201X}%
{Shell \MakeLowercase{\textit{et al.}}: Bare Demo of IEEEtran.cls for Computer Society Journals}
%

\maketitle
\begin{abstract}
Spatiotemporal Graph Neural Networks (ST-GNNs) and Transformers have shown significant promise in traffic forecasting by effectively modeling temporal and spatial correlations. However, rapid urbanization in recent years has led to dynamic shifts in traffic patterns and travel demand, posing major challenges for accurate long-term traffic prediction. The generalization capability of ST-GNNs in extended temporal scenarios and cross-city applications remains largely unexplored.
In this study, we evaluate state-of-the-art models on an extended traffic benchmark and observe substantial performance degradation in existing ST-GNNs over time, which we attribute to their limited inductive capabilities. Our analysis reveals that this degradation stems from an inability to adapt to evolving spatial relationships within urban environments. To address this limitation, we reconsider the design of adaptive embeddings and propose a Principal Component Analysis (PCA) embedding approach that enables models to adapt to new scenarios without retraining.
We incorporate PCA embeddings into existing ST-GNN and Transformer architectures, achieving marked improvements in performance. Notably, PCA embeddings allow for flexibility in graph structures between training and testing, enabling models trained on one city to perform zero-shot predictions on other cities. This adaptability demonstrates the potential of PCA embeddings in enhancing the robustness and generalization of spatiotemporal models. \textcolor{magenta}{\textit{The code is released in \href{https://github.com/Dreamzz5/PCA-Embedding}{code}.}}
\end{abstract}
	

\section{Introduction}

Recent advancements  in Spatiotemporal Graph Neural Networks (ST-GNNs) and Transformer models have opened up exciting possibilities for traffic prediction. These models excel at capturing both spatial and temporal dependencies in traffic data, offering promising results under stable conditions by leveraging the structure of transportation networks~\cite{GTS,DSTAGNN,PMMemNet,MegaCRN,CaST,liu2024largest,MemeSTN}. However, the fast-paced urbanization and constant change in modern cities pose unique challenges to accurate traffic predictions~\cite{zhang2020curb,CaST,zhou2023maintaining,ji2023self}. As cities grow, traffic patterns and demand shift unpredictably, requiring models that can keep up with these dynamics.

\begin{figure}[t]
	\centering
	\includegraphics[width=1\linewidth]{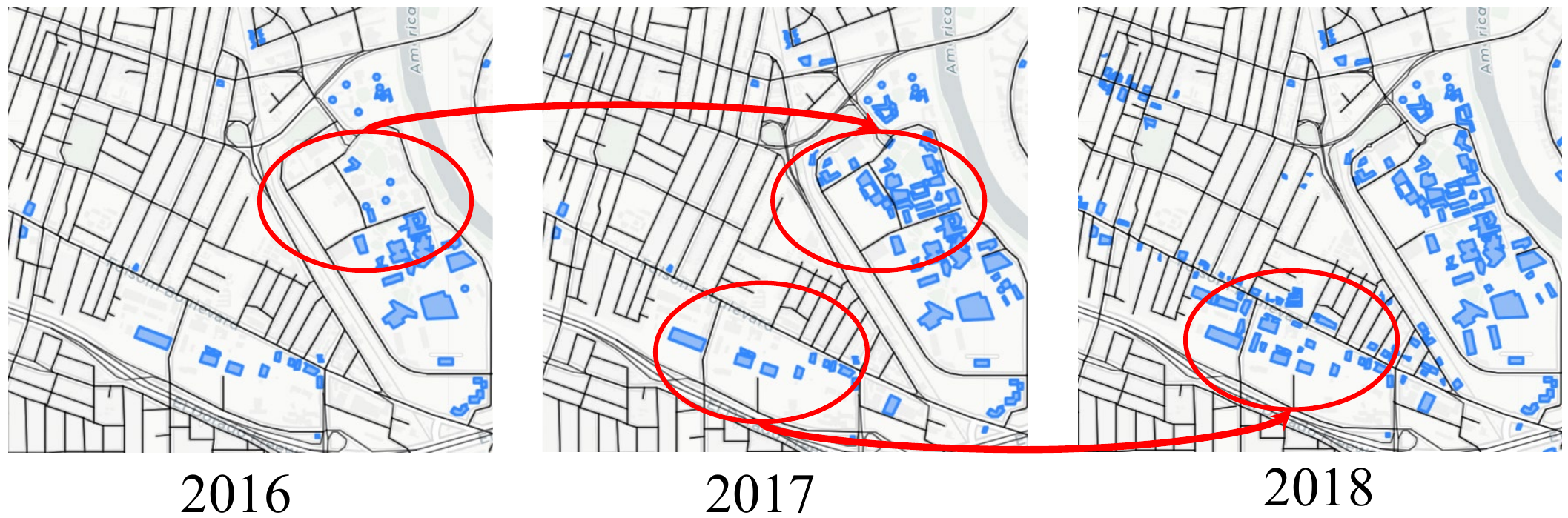} 
	\vspace{-12pt}
	\caption{\textbf{Spatiotemporal analysis of urban development patterns in Sacramento, California (2016-2018). The sequence demonstrates the progressive intensification of building density (shown in blue) and its implications for transportation demand modeling.} Red circles highlight key areas of urban transformation, indicating the dynamic nature of land use changes and their potential impact on travel demand patterns.}\label{fig:motivation}
	\vspace{-5pt}
\end{figure}

Figure~\ref{fig:motivation} illustrates the evolving dynamics of urban spatial morphology in Sacramento, California, from 2016 to 2018, highlighting a notable increase in building density within the study area. This spatial transformation significantly influences regional land use configurations and functional layouts. \textit{The emergence of new building clusters not only reshapes the urban landscape but also generates additional travel demand, thereby impacting existing transportation demand models. This fluidity in urban mobility poses a substantial challenge to current traffic prediction models, which typically assume stable spatial and travel patterns.}


\begin{figure*}[t]
	\centering
	\subfigure[PEMS03]{\includegraphics[width=0.48\linewidth]{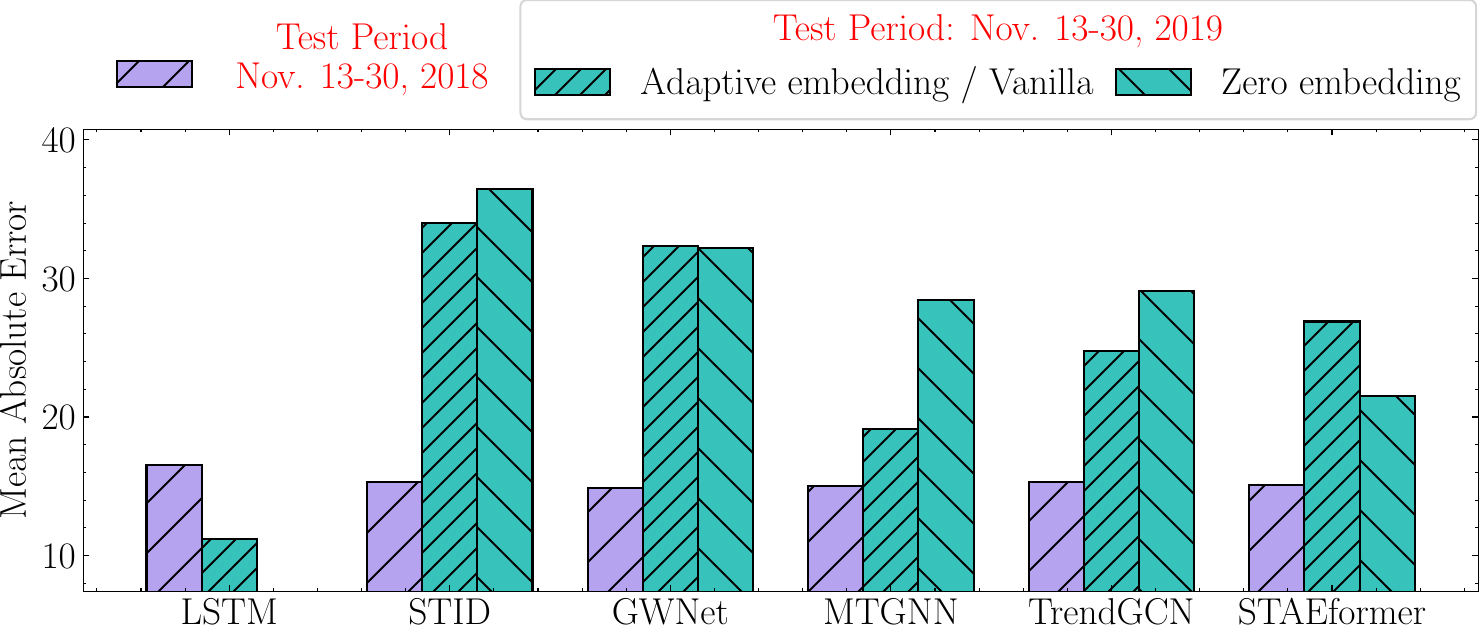}}
	\hfill
	\subfigure[PEMS04]{\includegraphics[width=0.48\linewidth]{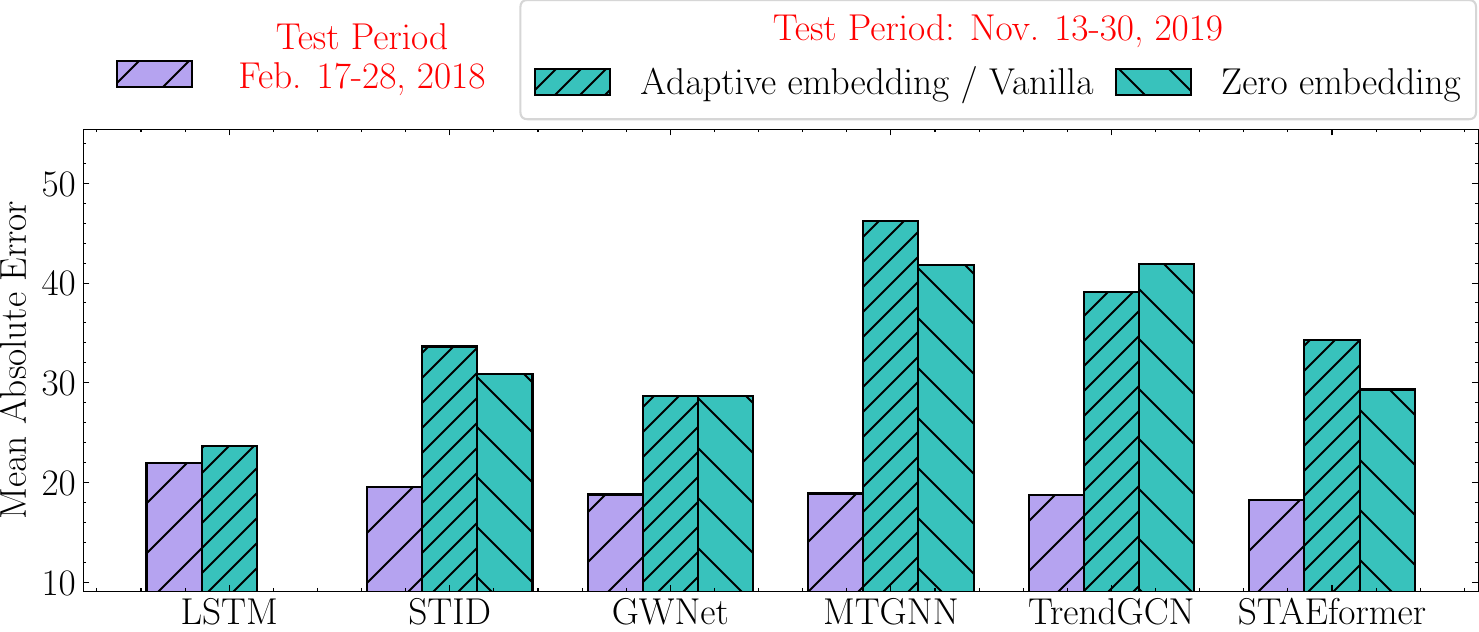}}
	
	\subfigure[PEMS07]{\includegraphics[width=0.48\linewidth]{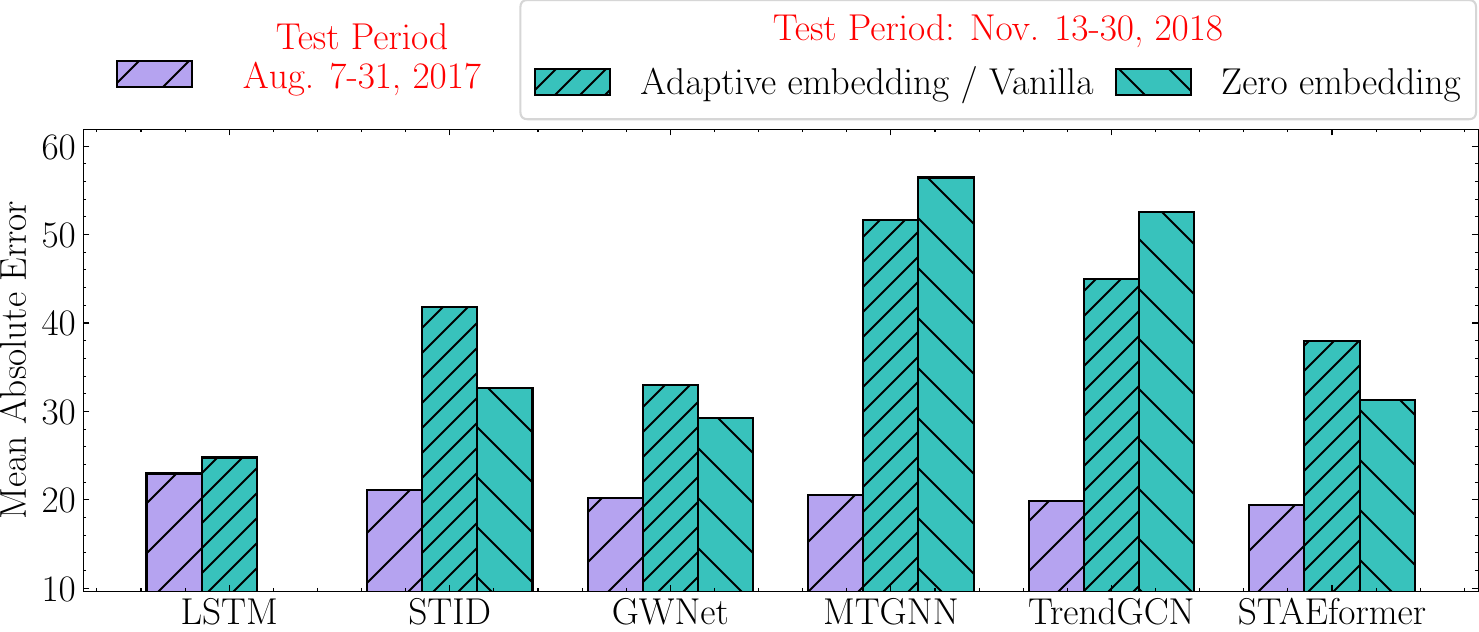}}
	\hfill
	\subfigure[PEMS08]{\includegraphics[width=0.48\linewidth]{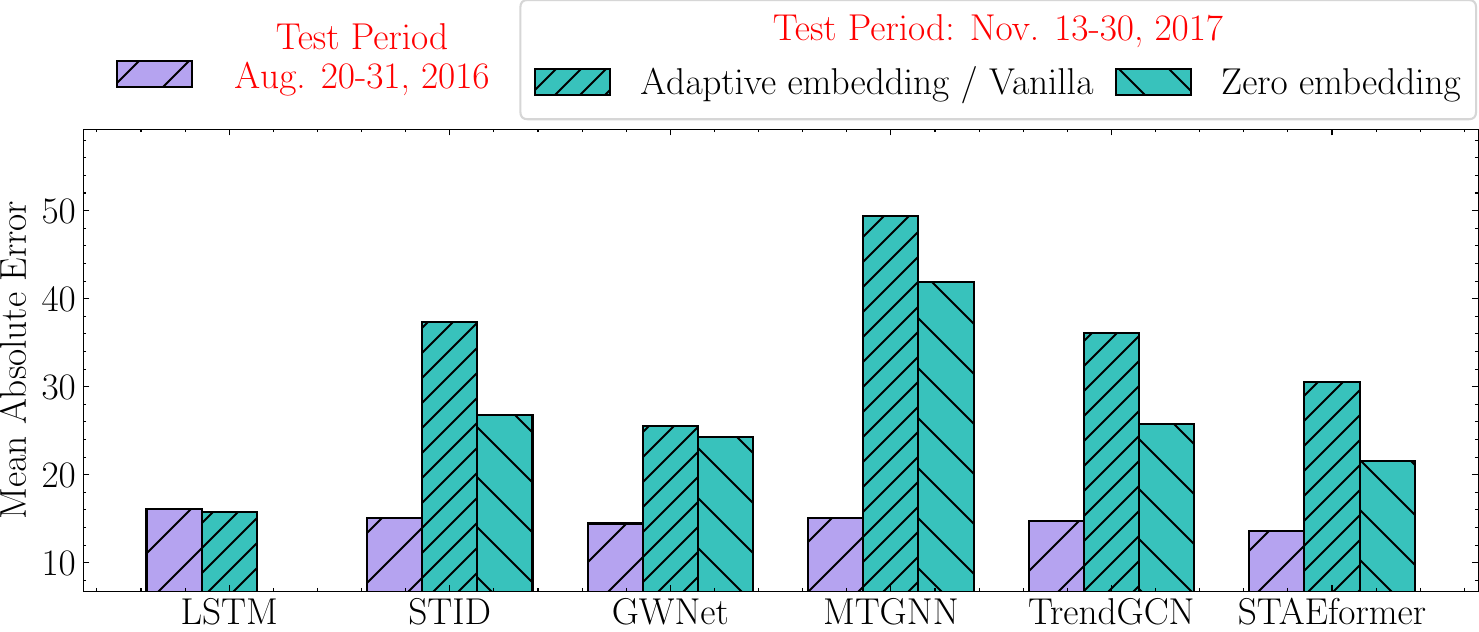}}
	
	\caption{\textbf{We conducted a comparative analysis of LSTM against state-of-the-art (SOTA) models, including , STID \cite{STID}, GWNet~\cite{GWNet}, AGCRN~\cite{AGCRN}, MTGNN~\cite{wu2020connecting}, TrendGCN~\cite{jiang2023enhancing}, and STAEformer~\cite{STAEformer}, assessing their performance on both in-distribution and out-of-distribution test datasets.} In the in-distribution scenario, we utilized the original dataset, which had only a few weeks' interval from the training set, while the out-of-distribution scenario employed our newly collected dataset with a one-year interval. For models employing adaptive embedding, we conducted tests using both the original adaptive embedding and a zero embedding strategy, in which the trained adaptive embeddings were set to zero to reduce potential bias. Our results showed that LSTM maintained consistent performance across both in-distribution and out-of-distribution scenarios. This finding strongly suggests that the observed performance degradation in other models is primarily attributable to shifts in spatial relationships. We further observed that using zero embedding to mitigate bias improved test performance, though optimal results were not achieved.}
	\label{fig:compare}
\end{figure*}

Traditional traffic prediction models have typically aimed to enhance performance in stable conditions~\cite{shao2023exploring,ASTGCN,DCRNN,DLTraff}, assuming relatively fixed spatial relationships between locations. However, our study challenges this assumption, showing that it often fails to hold in real-world cities undergoing continuous transformation. Infrastructure updates—such as new roads, station relocations, or the emergence of new urban centers—can drastically reshape spatial dynamics within a city, rendering previously learned spatial correlations outdated and diminishing the effectiveness of even advanced models.

To rigorously analyze this issue, we developed four long-term traffic benchmarks based on the California Department of Transportation (CalTrans) Performance Measurement System (PEMS)~\cite{chen2001freeway}. These benchmarks align with established standards~\cite{song2020spatial, guo2019attention}, using consistent sensors while gathering traffic data over multiple years. Our analysis uncovered a notable trend: while temporal patterns (e.g., weekday rush hours) remain relatively consistent, spatial relationships exhibit substantial variability over time. \figref{fig:compare} compares the performance of several models for traffic flow prediction, specifically LSTM, STID \cite{STID}, GWNet~\cite{GWNet}, AGCRN~\cite{AGCRN}, MTGNN~\cite{wu2020connecting}, TrendGCN~\cite{jiang2023enhancing}, and STAEformer~\cite{STAEformer}, across four datasets (PEMS03, PEMS04, PEMS07, and PEMS08) under two evaluation scenarios: in-distribution and out-of-distribution data.
Following \cite{darcet2023vision}, we introduce zero embeddings as a method to mitigate bias in our model. By setting biased embeddings to zero during testing, we evaluate the model's performance under unbiased conditions.

In \figref{fig:compare}, the purple bars represent the Mean Absolute Error (MAE) for in-distribution testing, which involves data from the same distribution as the training set. In contrast, the green bars show MAE for out-of-distribution testing, where the models are tested on data that may include distributional shifts. Out-of-distribution performance is further broken down by two strategies: adaptive (or "vanilla") embeddings (green bars with stripes) and zero embeddings (solid green bars), with the latter setting the learned adaptive embeddings to zero to potentially mitigate bias.
The results highlight that LSTM performs consistently across both in-distribution and out-of-distribution tests, with relatively low errors, indicating strong robustness to distributional shifts. However, other models exhibit greater variability, especially in the out-of-distribution case, where the zero embedding strategy often results in lower MAE compared to the adaptive embedding approach. This suggests that the zero embedding strategy may help control bias and improve generalization in scenarios where the test data distribution differs from training.

In this paper, we investigate the core limitations of current approaches in addressing spatial shifts within traffic prediction models. While methods like fine-tuning~\cite{zhuang2020comprehensive} and online learning~\cite{hoi2021online} show promise, the dynamic nature of cities complicates continuous updates, especially at scale. As a solution, we propose a training-free approach using PCA~\cite{wold1987principal}, enabling adaptation to spatial changes without necessitating model retraining, making it feasible for large-scale deployment.

In this paper, we investigate the fundamental limitations of current approaches to handling spatial shifts in traffic prediction models. While techniques such as fine-tuning~\cite{zhuang2020comprehensive} or online learning~\cite{hoi2021online} might seem promising, the constantly evolving nature of urban environments makes continuous model updates impractical, especially for large-scale deployments. Instead, we explore a training-free strategy based on PCA ~\cite{wold1987principal} that can effectively adapt to spatial shifts without requiring model retraining.

\textit{Additionally, we further expanded the definition of spatial shifts, which not only refers to different years within the same city but also to the model's capacity to be trained in city $A$ and generalize to city $B$ in a zero-shot manner.} We assessed the zero-shot generalization performance \cite{pourpanah2022review} of various models across different cities in the PEMS dataset, including both small-scale datasets \cite{guo2019attention} and largeST \cite{liu2024largest}.
\textbf{Key findings include}: 1) Models trained on larger-scale datasets exhibit stronger zero-shot generalization capabilities. 2) Transformer-based architectures outperform GNN-based models in terms of zero-shot generalization.
Our approach helps bridge the gap between disparate datasets, potentially advancing the development of future large-scale traffic models. 

\section{Problem Statements}
In traffic forecasting, we define a graph as \(\mathcal{G}=(\mathcal{V}, \mathcal{E}, A)\), where \(\mathcal{V}\) is the set of nodes, \(\mathcal{E} \subseteq \mathcal{V} \times \mathcal{V}\) represents the edges, and \(A\) is the adjacency matrix associated with the graph \(\mathcal{G}\). At each time step \(t\), the graph is associated with a dynamic feature matrix \(\mathbf{X}_t\) in the real-number space \(\mathbb{R}^{|\mathcal{V}| \times \mathcal{C}}\), where \(\mathcal{C}\) indicates the dimensionality of the node features (e.g., traffic volume, traffic speed, time of day and time of week).
Traffic forecasting involves developing and training a neural network model \(f_\theta\), formulated as: \(f_\theta: \left[X_t, A, E\right] \mapsto Y_t\), where $E$ presents the adaptive embedding layer learning from training data, \(X_t = \mathbf{X}_{(t-l_1): t}\) and \(Y_t = \mathbf{X}_{(t+1):(t+l_2)}\), with \(l_1\) and \(l_2\) representing the lengths of the input and output sequences, respectively.


\begin{figure*}[t]
	\centering
	\includegraphics[width=1\linewidth]{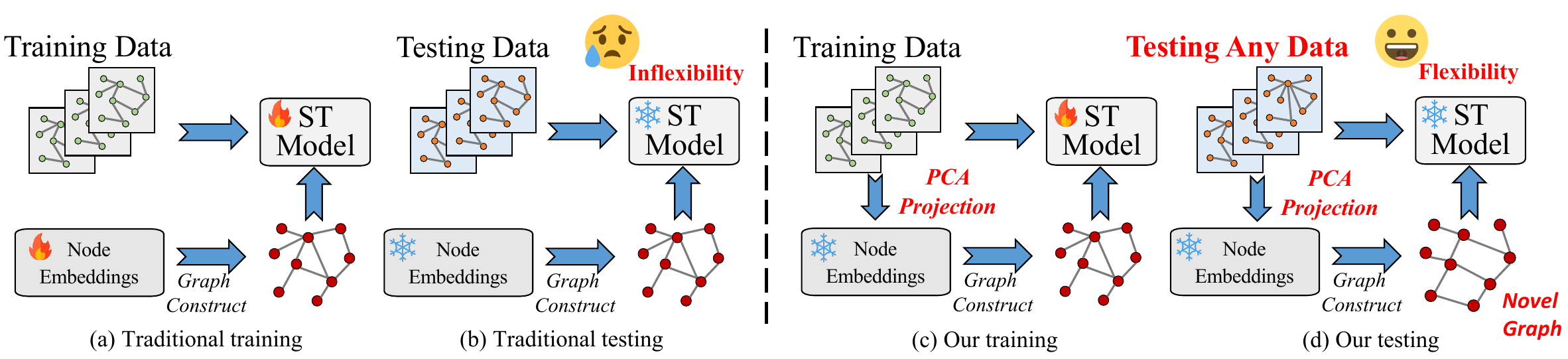}
	\caption{\textbf{We found that adaptive embedding restricts the model to perform inference on the same graph, which is unrealistic in dynamic traffic scenarios due to the continuous development of cities.} We propose a novel testing-time adaptive strategy that requires no additional training, where 'fire' indicates the optimization of target parameters and 'snowflake' represents the freezing of model parameters. Specifically, we apply PCA to compress the training input representations into a low-dimensional space. During testing, the same projection matrix is used to map the input into the same space. Within this new projected embedding, we can construct an entirely new graph structure, enabling the model to adapt to novel spatial relationships.}
	\label{fig:framework}
\end{figure*}

\begin{figure}[h]
	\centering
	\includegraphics[width=1\linewidth]{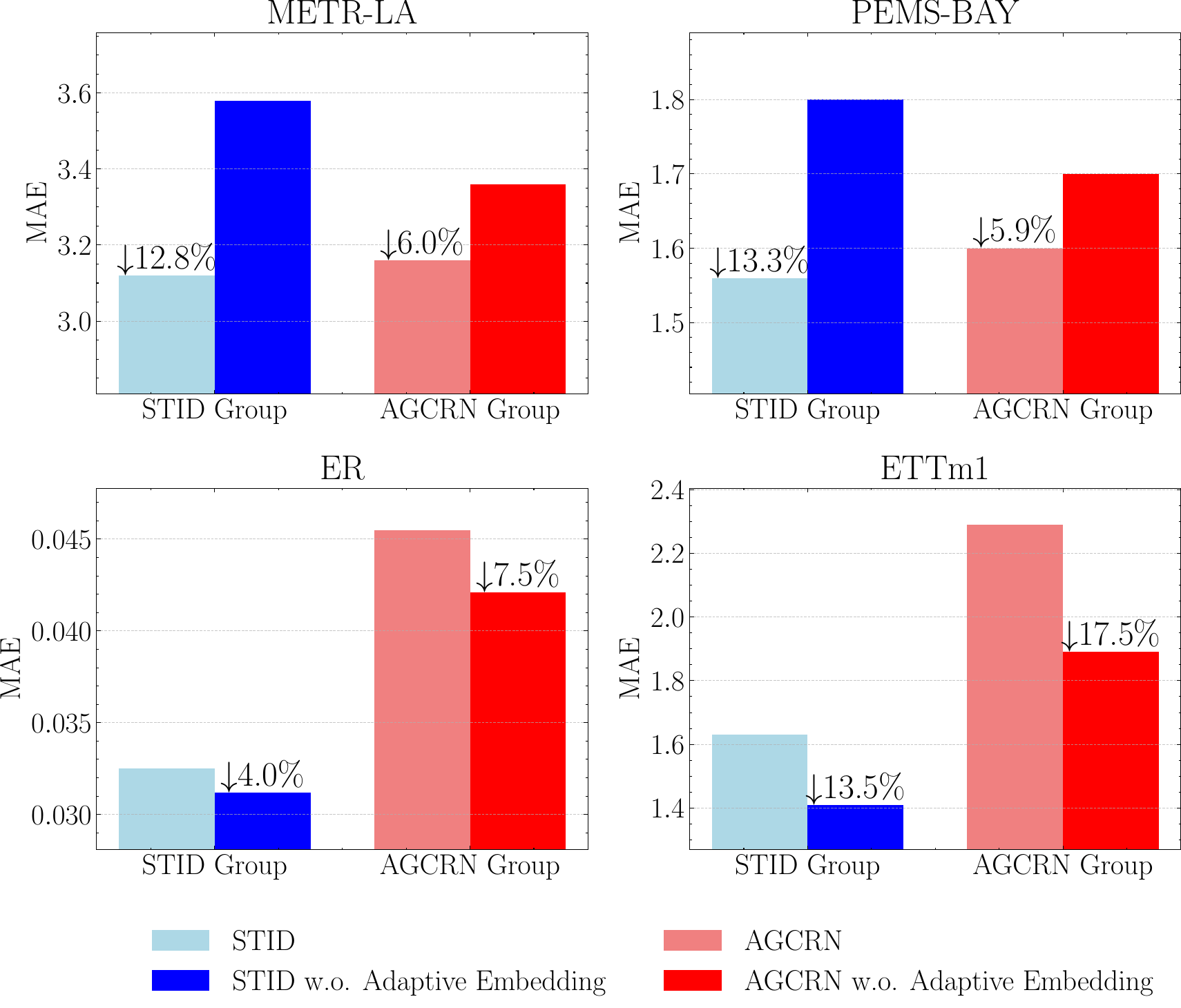}
	\vspace{-15pt}
	\caption{\textbf{Performance comparison of STID and AGCRN models with and without adaptive embedding across four datasets (METR-LA, PEMS-BAY, ER, and ETTm1).} We observe that employing trainable adaptive embeddings results in excessive spatial distinguishability, leading to a performance degradation of the model compared to those without adaptive embeddings.}
	\label{fig:overdistinguishability}
\end{figure}

\section{Methodology}

\subsection{Overview of Adaptive Embedding Layer}

In traffic forecasting \cite{shao2023exploring}, spatial indistinguishability poses a significant challenge, where time series with closely aligned historical patterns over specific observation windows display substantial divergence in their future trajectories. To address this issue, researchers have leveraged Tobler's first law of geography \cite{tobler1970computer}, by introducing GNNs into traffic prediction to resolve spatial indistinguishability. 

Message passing in GNNs operates on the principle of local similarity, where proximate nodes are expected to demonstrate similar traffic patterns \cite{luo2023stg4traffic}. Spatial-temporal graphs for traffic prediction typically use either road connection distances \cite{STGCN} or absolute physical coordinates \cite{DCRNN} to calculate edge weights. However, connectivity relationships in these predefined graphs are often incomplete or biased, as they rely heavily on supplementary data and human expertise, which complicates the task of capturing a comprehensive panorama of spatial dependencies.

To address this challenge, the adaptive graph approach is proposed to leverage parameter representations of adaptive embeddings, which are continuously updated throughout the training phase to minimize model errors. Adaptive graph aims to identify biases stemming from human-defined concepts and to uncover hidden spatial dependencies within the data. Mainstream adaptive graph learning methods utilize randomly initialized learnable matrices~\cite{shao2022decoupled,GWNet,wu2020connecting,ye2021coupled,han2021dynamic,sun2022spatial,lu2020spatiotemporal,zhang2022adapgl}. Traditionally, the adaptive graph \cite{AGCRN} is generated as follows:
\begin{equation}
	\tilde{\boldsymbol{A}}_{\text{adp}} = \operatorname{SoftMax}\left(\operatorname{ReLU}\left(E E^T\right)\right),
\end{equation}
where $E\in \mathbb{R}^{N\times C}$ is adaptive embeddings, initialized randomly, with $N$ representing the number of nodes and $C$ the embedding dimension. 

Recently, more innovative approaches have been introduced that bypass the construction of an adaptive graph, instead using learnable node embedding techniques. Models such as STID \cite{STID} and STAEformer \cite{STAEformer} introduce trainable adaptive embeddings, thereby enhancing the model’s capability to distinguish among similar historical patterns. Each of these strategies offers a distinct approach to addressing the fundamental challenge of spatial indistinguishability in multivariate time series forecasting.

\subsection{The Limitation of Adaptive Embedding}
\begin{figure*}[t]
	\centering
	
	\includegraphics[width=\linewidth]{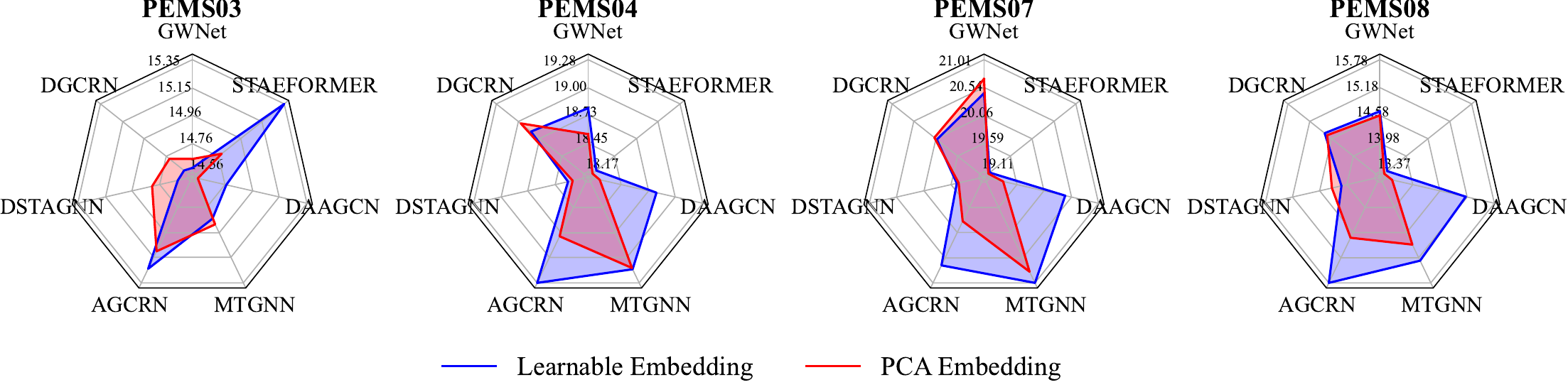}
	\vspace{-15pt}
	\caption{\textbf{We conducted a comparative analysis of model performance on the PEMS benchmark using PCA embeddings versus learnable embeddings.} The MAE results reveal that employing PCA (frozen) embeddings does not deteriorate training outcomes. Surprisingly, in certain instances, it leads to substantially improved performance by mitigating over-fit issue.}
	\label{fig:fix_perf}
\end{figure*}

In this section, we delve into three key limitations inherent to Adaptive Embedding: excessive spatial indistinguishability, lack of inductive capacity, and limited transferability.
\begin{itemize}[leftmargin=*]
	\item[$\bullet$] \textbf{Lack of Inductive Capacity:} As urban environments evolve, new infrastructure and traffic patterns emerge, rendering previously indistinguishable locations potentially distinct over time. However, Adaptive Embedding, relying on fixed embeddings during inference, is inherently limited in its ability to adapt to such dynamic changes. This constraint is visually depicted in  \figref{fig:fix_perf}, where the model's performance diminishes as the environment diverges from its training distribution. \figref{fig:fix_perf} reveals that while the LSTM model maintains relatively low MAE values across in-distribution and out-of-distribution settings, other SOTA models (GWNet, AGCRN, MTGNN, TrendGCN, STAEformer) exhibit a pronounced sensitivity to distributional shifts, highlighting the fact that  adaptive embedding lack of inductive capacity.
	
	\item[$\bullet$] \textbf{Excessive Spatial Distinguishability:} As described in \cite{shao2023exploring}, adaptive embeddings may be affected by excessive spatial distinguishability, particularly for datasets where spatial relationships are less critical. Empirical analysis presented in \figref{fig:overdistinguishability} indicates that datasets with high spatial indistinguishability (such as METR-LA and PEMS-BAY) benefit significantly from trainable adaptive embeddings. However, datasets with low spatial indistinguishability (such as ExchangeRate and ETTm1) experience performance degradation in models like STID and AGCRN when such components are introduced. \cite{shao2023exploring} observe that spatially indistinguishable samples constitute only a very small portion (approximately one in a thousand) of total observations in datasets requiring spatial differentiation, leading to a risk of overfitting when using adaptive embeddings. Subsequent experiments in \figref{fig:fix_perf} reveal that, compared to PCA embeddings, trainable adaptive embeddings may induce over-distinguishability on datasets where spatial relationships are important, resulting in overfitting to the training data and weakened model generalization capability.
	\item[$\bullet$] \textbf{Limited Transferability:}  A critical limitation of Adaptive Embedding lies in its restricted transferability across different scenarios and deployments. As sensor deployments evolve over time, maintaining a fixed graph size becomes increasingly challenging \cite{chen2001freeway}, especially when urban environments require modifications to their sensor networks through additions, decommissioning, or temporary failures \cite{choe2002freeway}. The inherent design of adaptive embeddings, being tightly coupled with specific sensor, severely limits their applicability across different cities \cite{wu2020comprehensive}, requiring complete model retraining for each new deployment \cite{yuan2024unist}. This inflexibility is particularly problematic in rapidly developing urban areas where infrastructure changes are frequent \cite{wang2024evaluating}, leading to significant computational overhead and resource requirements \cite{liu2024largest}. While recent research has proposed potential solutions, such as meta-learning frameworks \cite{pan2019urban} and transfer learning strategies \cite{wang2018cross}, these approaches fail to address the core issue of adaptive embedding. Consequently, most mainstream ST-models lack generalizability.
\end{itemize} 

\begin{table*}[t]
	\centering
	\caption{ST-model's  zero-shot performance with PCA-embedding. We use $A \rightarrow B$ to indicate that the model is trained on the dataset $A$  and tested on the dataset $B$. }
	\label{tab:trans}
	\renewcommand\arraystretch{1.3}
	\begin{sc}
		\resizebox{\linewidth}{!}{%
			\begin{tabular}{|c|ccc|ccc|ccc|ccc|}
				\toprule
				\multirow{2}{*}{Model} & \multicolumn{3}{c|}{PEMS07  $\rightarrow$ PEMS04} & \multicolumn{3}{c|}{PEMS03 $\rightarrow$ PEMS04}  & \multicolumn{3}{c|}{PEMS07 $\rightarrow$ PEMS08} & \multicolumn{3}{c|}{PEMS03 $\rightarrow$ PEMS08}\\
				\cline{2-13}
				& MAE & RMSE & MAPE & MAE & RMSE & MAPE & MAE & RMSE & MAPE & MAE & RMSE & MAPE\\
				\midrule
				DGCRN & 28.52& 42.86& 23.77\% & 32.68 & 47.95 & 26.32\%  & 21.85 & 32.96 & 15.53\% & 26.32& 38.18& 34.43 \% \\
				STID & 25.32 & 38.32 & 18.85\% & 32.70& 48.58& 26.53\% & 19.93 & 29.89 & 13.63\% & 26.88& 39.04& 32.01\% \\
				D$^2$STGNN & 31.61 & 47.47 & 23.28\% & 35.01& 49.57& 28.43\% & 21.41 & 32.13 & 13.37\% & 31.10& 42.74& 35.68\% \\
				GWNet & 26.05 & 39.07 &  19.02\% & 28.47& 42.99& 23.44\% &   18.81 & 28.79 &  12.02\% & 23.84& 35.00& 29.67\% \\
				STAEformer & \bf \cellcolor{green!30} 25.20 & \bf \cellcolor{green!30} 37.96 &\bf \cellcolor{green!30} 18.90\% &\bf \cellcolor{green!30} 27.97& \bf \cellcolor{green!30} 25.72&\bf \cellcolor{green!30} 22.22\% & \bf \cellcolor{green!30} 18.44 & \bf \cellcolor{green!30} 28.53 &\bf \cellcolor{green!30}  11.79\% & \bf \cellcolor{green!30} 23.58&\bf \cellcolor{green!30} 31.88&\bf \cellcolor{green!30}  28.37\% \\ 
				\bottomrule
				\toprule
				\multirow{2}{*}{Model}  & \multicolumn{3}{c|}{{\small San Diego  $\rightarrow$ Los Angeles}} & \multicolumn{3}{c|}{{\small San Diego $\rightarrow$ Bay Area}} & \multicolumn{3}{c|}{{\small Bay Area  $\rightarrow$ San Diego}} & \multicolumn{3}{c|}{{\small Bay Area $\rightarrow$ Los Angeles}} \\
				\cline{2-13}
				& MAE & RMSE & MAPE & MAE & RMSE & MAPE & MAE & RMSE & MAPE & MAE & RMSE & MAPE\\
				\midrule
				DGCRN & 27.80 & 41.90 & 20.50\% & 27.62 & 42.34 & 21.30\% & 25.42 & 38.85 & 17.20\% & 25.98 & 40.05 & 16.40\% \\
				STID & 25.82 & 39.56 & 17.57\% & 25.65 & 39.97 & 18.32\% & 23.61 & 36.11 & 15.99\% & 24.15 & 37.26 & 15.27\% \\
				D$^2$STGNN & 30.75 & 46.20 & 22.65\% & 30.55 & 46.70 & 23.55\% & 28.10 & 42.90 & 19.00\% & 28.72 & 44.20 & 18.10\% \\
				GWNet & 25.61 & 38.64 & 20.20\% & 25.53 & 39.26 & 20.31\% & 23.42 & 35.78 & 18.50\% & 23.95 & 36.90 & 17.60\% \\
				STAEformer & \bf \cellcolor{green!30} 24.95 & \bf \cellcolor{green!30} 37.50 & \bf \cellcolor{green!30} 17.10\% & \bf \cellcolor{green!30} 24.80 & \bf \cellcolor{green!30} 37.80 & \bf \cellcolor{green!30} 17.80\% & \bf \cellcolor{green!30} 22.80 & \bf \cellcolor{green!30} 35.00 & \bf \cellcolor{green!30} 15.50\% & \bf \cellcolor{green!30} 23.30 & \bf \cellcolor{green!30} 36.10 & \bf \cellcolor{green!30} 14.80\% \\
				\bottomrule
			\end{tabular}%
		}
	\end{sc}
\end{table*}

\subsection{Principal Component Analysis Embedding}
To address the aforementioned limitations of adaptive embedding, we propose PCA embedding as an alternative approach. PCA embedding effectively mitigates the three key challenges through its statistical and data-driven nature, while maintaining the ability to capture essential spatiotemporal relationships.  Formally, due to the inherent periodicity in traffic data, we first divide each day into equal time slots to obtain $Z \in \mathbb{R}^{D \times N \times T}$, where $D$ is the number of days, $N$ represents the number of nodes, and $T$ is the number of time slots in a day (e.g., with a 5-minute sampling interval, $T=288$). We then apply PCA to obtain the embedding matrix for each day:
$
E_{pca}^d = Z^d \cdot \mathbf{P} \in \mathbb{R}^{N \times C}, \ d \in \{1,...,D\},
$
where $\mathbf{P}$ is the projection matrix generated from PCA, and $C$ is the dimension of PCA embedding. Subsequently, we average the PCA embeddings across all training days to obtain the final node representations:
$$
E_{pca}^{train} = \frac{1}{D} \sum_{d=1}^{D} E_{pca}^d \in \mathbb{R}^{N \times C}
$$
In the testing phase, the same PCA projection matrix, \(\mathbf{P}\), is utilized to ensure consistency in feature extraction. 
\begin{equation}
	E_{pca}^{test} = Z^{val} \cdot \mathbf{P}.
\end{equation}
To mitigate information leakage, a small subset (5\%) of the validation set is designated as the validation subset. PCA is then applied to distill the key features that capture the system's spatiotemporal dynamics.

\subsubsection{Advantage of PCA Embedding}
The PCA embedding method effectively overcomes three key constraints of adaptive embeddings through its rigorous statistical basis and data-driven adaptability.  Below, we detail how PCA embedding addresses each limitation:

Firstly, in terms of limited inductive capacity, PCA's generalization capabilities are notable, as it captures essential statistical features of the input data rather than depending on fixed, trainable parameters. As environmental dynamics shift, PCA inherently adapts to distribution changes by recalculating features from updated data inputs, thereby preserving its performance across varying conditions. This contrasts with static embeddings that may require retraining to adapt.

Secondly, PCA mitigates the challenge of excessive spatial distinguishability via its orthogonal basis representation, ensuring that extracted spatial features remain mutually independent. By appropriately selecting a subset of principal components that explain a sufficient fraction of the total variance (e.g., $\sum_{i=1}^k \lambda_i / \sum_{i=1}^n \lambda_i \geq \theta$), PCA maintains a balanced spatial representation without overfitting the training data. This characteristic makes it especially robust for scenarios requiring spatial generalization. For experimental comparison, please refer to \figref{fig:fix_perf} and the detailed description in the experimental part.

Thirdly, the limitation of poor transferability is addressed effectively by PCA's inherent statistical methodology, which allows for seamless application across different scenarios and deployments. In cases where sensor networks undergo alterations (such as the addition, decommissioning, or failure of nodes), only the PCA-derived features need recalculating ($\mathbf{X}_{\text{new}} \xrightarrow{\text{PCA}} E_{\text{adapted}}$), eliminating the necessity to retrain an entire model. Our experiments, summarized in \tableref{tab:trans}, demonstrate the zero-shot capabilities of the spatiotemporal model, particularly in both small-scale and large-scale settings. These results highlight the significant potential of the Transformer architecture.

Overall, PCA's capacity to extract critical ST relationships while offering enhanced generalization, controlled spatial discrimination, and improved cross-context transferability renders it an essential tool for evolving urban contexts. We summarize the advantages of PCA embedding below:
\begin{equation*}
	\begin{aligned}
		\begin{cases}
			\text{Inductive Capacity: } \quad \mathbf{X}_{\text{new}} \xrightarrow{\text{PCA}} E_{\text{adapted}} \\
			\text{Balanced Distinguishability: } \quad \sum_{i=1}^k \lambda_i \\
			\text{Transferability: } \quad \mathbf{X}_{\text{city}_1} \xrightarrow{\text{PCA}} E_{\text{city}_1}, \mathbf{X}_{\text{city}_2} \xrightarrow{\text{PCA}} \mathbf{E}_{\text{city}_2}
		\end{cases}
	\end{aligned}
\end{equation*}

\begin{table}[h]
	\small
	\centering
	\caption{ Information of the cross-year datasets.}\label{tab:detail}
	\begin{tabular}{ccccc}
		\hline    Datasets & Nodes & Edges & Year&  Time Range  \\
		\hline 
		PEMS03 & 358 & 547 & 2018/2019 &09/01 - 11/30  \\
		PEMS04 & 307 & 340 &2018/2019& 01/01 - 02/28 \\
		PEMS07 & 883 & 866 &2017/2018& 05/01 - 08/31\\
		PEMS08 & 170 & 295 &2016/2017&  07/01 - 08/31 \\
		\hline
	\end{tabular}
\end{table}

\section{Experiment}
In this section, we analyze the performance of PCA-embedding by addressing the following research questions:
\begin{itemize}[leftmargin=*,itemsep=0em,parsep=0.0em] 
	\item[$\bullet$] \textbf{RQ1}: Is the performance of PCA-embedding consistent with Adaptive-embedding?
	\item[$\bullet$] \textbf{RQ2}: How does PCA-embedding perform in zero-shot generalization?
	\item[$\bullet$] \textbf{RQ3}: How does PCA-embedding perform when facing spatial shifts?
	\item[$\bullet$] \textbf{RQ4}: Why PCA-embedding work in zero-shot generalization?
	\item[$\bullet$] \textbf{RQ5}: How does PCA-embedding perform compared to fine-tuning methods?

\end{itemize}
To evaluate the spatial shift performance of current ST-GNNs, we introduced four traffic benchmark datasets: PEMS03-2019, PEMS04-2019, PEMS07-2018, and PEMS08-2017, which align with existing standards \cite{song2020spatial, guo2019attention} and utilize the same sensors to capture traffic data across different years. The detail information of datasets is shown in \tableref{tab:detail}. The experimental setup involved predicting the next 12 steps based on the preceding 12 steps of historical data \cite{DCRNN, GWNet}.  Following the methodology described in \cite{guo2019attention}, we chronologically split the data into training, validation, and test sets, maintaining a 6:2:2 ratio for both same and cross year data. \textit{To prevent information leakage, we utilized 5\% of the data from the second year for PCA dimensionality reduction to generate embeddings, and then tested on the remaining 95\%.}

\noindent\textbf{Training Detail.}
Our experiments were conducted on a GPU server equipped with eight GeForce GTX A100 graphics cards, utilizing the PyTorch 1.13.1 framework. The raw time-series data were standardized using z-score normalization \cite{cheadle2003analysis}. Training was terminated early if the validation error stabilized within 15-20 epochs or showed no improvement after 200 epochs, with the best model retained based on validation performance \cite{luo2023dynamic}. We strictly adhered to the model parameters and settings outlined in the original papers and performed multiple rounds of parameter tuning to achieve optimal results. 
Model performance was assessed using Mask-Based Root Mean Square Error (RMSE), Mean Absolute Error (MAE), and Mean Absolute Percentage Error (MAPE), with zero values excluded as they represent noisy data.

\begin{table}[t]
	\caption{Comparison with zero-embedding and fine-tune.}\label{tab:res2}
	\centering
	\renewcommand\arraystretch{1.25}
	\begin{sc}
		\begin{adjustbox}{width=0.45\textwidth}
			\setlength{\tabcolsep}{0.9mm}{\begin{tabular}{c|p{25mm}ccc}
					\hline Model &  Strategy &  MAE  & RMSE &  MAPE(\%) \\
					\hline 			
					\multirowcell{4}{ PEMS03 \\  2018 \\ $\downarrow$ \\ 2019}& AGCRN&  
					20.94&31.85&20.16\% \\
					&+Fine-tune & 18.08 & 29.01 & 19.05\% \\
					\cline{2-5}
					&+Zero-emb & 19.99 & 30.19 &19.85\%\\
					& \bf \cellcolor{green!30} +PCA-emb &\bf \cellcolor{green!30}  18.53&\bf \cellcolor{green!30}29.47&\bf \cellcolor{green!30}19.57\%\\
					
					\hline
					\multirowcell{4}{ PEMS04 \\  2018 \\ $\downarrow$ \\ 2019}& GWNet&  48.08&74.09&40.29\%\\
					&+Fine-tune & 23.60 &37.67 &19.17\% \\
					\cline{2-5}
					&+Zero-emb&  29.75&48.17&20.56\%\\
					& \bf \cellcolor{green!30} +PCA-emb &\bf \cellcolor{green!30}23.89&\bf \cellcolor{green!30} 38.47 &\bf \cellcolor{green!30}19.87\% \\
					\hline
					\multirowcell{4}{ PEMS07 \\  2017 \\ $\downarrow$ \\ 2018}& STID
					&38.96&55.61&41.11\%  \\
					&+Fine-tune& 23.66 & 35.45 & 13.26\% \\
					\cline{2-5}
					&+Zero-emb& 32.23 & 48.44 & 15.72\% \\
					& \bf \cellcolor{green!30} +PCA-emb &\bf \cellcolor{green!30} 24.29&\bf \cellcolor{green!30} 36.88&\bf \cellcolor{green!30} 13.40\% \\
					\hline
					\multirowcell{4}{ PEMS08 \\  2016 \\ $\downarrow$ \\ 2017}& 
					STAEformer&  33.45 &49.53 &35.34\% \\
					&+Fine-tune &14.42 &25.34 &9.02\%  \\
					\cline{2-5}
					&+Zero-emb&  24.09 &38.98 & 16.21\% \\
					& \bf \cellcolor{green!30} +PCA-emb &\bf \cellcolor{green!30} 14.70 &\bf \cellcolor{green!30}25.67 &\bf \cellcolor{green!30}9.10\%  \\
					\hline
			\end{tabular}}
		\end{adjustbox}
	\end{sc}
\end{table}
\begin{table*}[t]
	\centering
	\small
	\renewcommand{\arraystretch}{1.3}
	\tabcolsep=1mm
	\caption{Performance comparisons are presented, with the best-performing baseline results highlighted in bold.   We use $A \rightarrow B$ to indicate that the model is trained on the $A$ year’s data and tested on the $B$ year’s data.}\label{tab:performance1}
	\begin{sc}
		\resizebox{\textwidth}{!}{
			\begin{tabular}{lcccc|ccc|ccc|ccc}
				\toprule
				\multirow{2}{*}{Data} & \multirow{2}{*}{Method} & \multicolumn{3}{c}{Horizon 3} & \multicolumn{3}{c}{Horizon 6} & \multicolumn{3}{c}{Horizon 12} & \multicolumn{3}{c}{Average} \\ \cline{3-14} 
				&  & MAE & RMSE & MAPE & MAE & RMSE & MAPE & MAE & RMSE & MAPE & MAE & RMSE & MAPE \\ 
				\hline \hline
				\multirowcell{13}{\bf PEMS03 \\ \bf 2018 \\ $\downarrow$ \\ \bf 2019} &  AGCRN  & 24.22 & 30.78 & 753.18\% & 28.51 & 36.01 & 794.16\% & 32.87 & 42.43 & 933.50\% & 28.04 & 35.78 & 820.77\% \\
				& \bf +PCA  & \bf 15.37 & \bf 20.19 & \bf 52.82\% & \bf 17.47 & \bf 23.33 & \bf 55.55\% & \bf 21.86 & \bf 30.14 & \bf 64.75\% & \bf 17.94 & \bf 24.07 & \bf 56.79\% \\\cline{2-14}
				& TrendGCN  & 28.99 & 33.78 & 1086.93\% & 22.60 & 27.61 & 803.78\% & 24.08 & 29.97 & 834.69\% & 24.74 & 29.86 & 899.05\% \\
				& \bf +PCA  & \bf 14.27 & \bf 18.54 & \bf 60.79\% & \bf 15.98 & \bf 21.52 & \bf 60.72\% & \bf 21.07 & \bf 29.68 & \bf 75.32\% & \bf 16.55 & \bf 22.33 & \bf 64.59\% \\\cline{2-14}
				& STID  & 24.58 & 33.71 & 503.27\% & 34.22 & 47.80 & 676.26\% & 49.16 & 66.35 & 1001.85\% & 33.99 & 46.70 & 686.25\% \\
				& \bf +PCA  & \bf 7.86 & \bf 11.62 & \bf 19.95\% & \bf 9.10 & \bf 13.67 & \bf 23.09\% & \bf 11.16 & \bf 16.86 & \bf 27.93\% & \bf 9.18 & \bf 13.75 & \bf 23.16\% \\\cline{2-14}
				 & MTGNN  & 16.62 & 22.05 & 62.66\% & 20.00 & 27.28 & 66.18\% & 28.29 & 40.58 & 81.99\% & 20.78 & 28.66 & 68.25\% \\
				& \bf +PCA  & \bf 9.00 & \bf 13.66 & \bf 23.35\% & \bf 10.67 & \bf 17.09 & \bf 26.83\% & \bf 13.93 & \bf 21.57 & \bf 32.48\% & \bf 10.93 & \bf 17.06 & \bf 27.08\% \\\cline{2-14}
				& GWNET  & 31.05 & 36.36 & 1157.81\% & 31.15 & 36.85 & 871.96\% & 38.50 & 45.85 & 1049.89\% & 32.33 & 37.97 & 966.58\% \\
				& \bf +PCA  & \bf 15.00 & \bf 20.24 & \bf 39.98\% & \bf 19.86 & \bf 26.63 & \bf 45.84\% & \bf 24.55 & \bf 33.14 & \bf 53.97\% & \bf 19.01 & \bf 25.66 & \bf 44.25\% \\\cline{2-14}
				& DGCRN  & 42.31 & 51.09 & 1093.60\% & 30.94 & 37.05 & 820.72\% & 34.54 & 42.04 & 904.42\% & 33.35 & 40.36 & 862.39\% \\
				& \bf +PCA  & \bf 11.77 & \bf 16.57 & \bf 50.01\% & \bf 12.00 & \bf 17.07 & \bf 40.94\% & \bf 15.48 & \bf 22.33 & \bf 46.10\% & \bf 12.63 & \bf 18.00 & \bf 44.39\% \\\cline{2-14}
				& STAEFORMER  & 20.51 & 22.56 & 890.56\% & 26.53 & 29.20 & 1077.78\% & 36.38 & 40.67 & 1329.79\% & 26.89 & 29.74 & 1073.63\% \\
				& \bf +PCA  & \bf 9.89 & \bf 13.68 & \bf 35.02\% & \bf 11.11 & \bf 15.63 & \bf 40.67\% & \bf 13.73 & \bf 19.12 & \bf 52.65\% & \bf 11.31 & \bf 15.79 & \bf 41.87\% \\\cline{2-14}
				& D$^2$STGNN  & 10.02 & 14.32  & 334.34\%  & 14.45 & 20.38 & 417.47\% & 19.64 & 28.37 & 475.09\% & 14.20 & 20.19 & 393.09\% \\
				& \bf +PCA  & \bf 9.08 & \bf 13.36 & \bf 20.26\% & \bf 10.71 & \bf 15.66 & \bf 23.20\% & \bf 13.52 & \bf 19.76 & \bf 26.61\% & \bf 10.58 & \bf 15.56 & \bf 24.12\% \\ \bottomrule
				\cline{2-14}
				& \cellcolor{green!30} Improve 
				& \cellcolor{green!30}$\textbf{48.3\%}\downarrow$ 
				& \cellcolor{green!30}$\textbf{46.2\%}\downarrow$ 
				& \cellcolor{green!30}$\textbf{44.9\%}\downarrow$ 
				& \cellcolor{green!30}$\textbf{42.6\%}\downarrow$ 
				& \cellcolor{green!30}$\textbf{39.7\%}\downarrow$ 
				& \cellcolor{green!30}$\textbf{38.7\%}\downarrow$ 
				& \cellcolor{green!30}$\textbf{91.0\%}\downarrow$ 
				& \cellcolor{green!30}$\textbf{90.2\%}\downarrow$ 
				& \cellcolor{green!30}$\textbf{90.2\%}\downarrow$ 
				& \cellcolor{green!30}$\textbf{46.5\%}\downarrow$ 
				& \cellcolor{green!30}$\textbf{40.3\%}\downarrow$ 
				& \cellcolor{green!30}$\textbf{90.5\%}\downarrow$ \\

				\hline\hline
				\multirowcell{13}{\bf PEMS04 \\ \bf 2018 \\ $\downarrow$ \\ \bf 2019} &  AGCRN  & 39.62 & 59.95 & 34.77\% & 48.85 & 73.62 & 40.49\% & 58.81 & 87.95 & 49.23\% & 48.11 & 72.46 & 40.44\% \\
				& \bf +PCA  & \bf 26.22 & \bf 40.16 & \bf 22.91\% & \bf 30.57 & \bf 46.48 & \bf 26.55\% & \bf 40.30 & \bf 59.98 & \bf 38.40\% & \bf 31.57 & \bf 47.75 & \bf 28.45\% \\\cline{2-14}
				& TrendGCN  & 32.88 & 50.75 & 29.74\% & 38.69 & 59.39 & 33.68\% & 47.72 & 71.79 & 39.97\% & 39.08 & 59.69 & 33.96\% \\
				& \bf +PCA  & \bf 24.44 & \bf 38.72 & \bf 23.23\% & \bf 29.42 & \bf 46.49 & \bf 27.05\% & \bf 37.61 & \bf 58.71 & \bf 33.18\% & \bf 29.79 & \bf 46.95 & \bf 27.26\% \\\cline{2-14}
				& STID  & 25.72 & 41.44 & 19.98\% & 33.04 & 54.41 & 25.33\% & 45.24 & 74.69 & 35.21\% & 33.64 & 55.25 & 25.98\% \\
				& \bf +PCA  & \bf 20.76 & \bf 34.17 & \bf 16.13\% & \bf 24.29 & \bf 39.53 & \bf 19.61\% & \bf 31.07 & \bf 49.30 & \bf 26.05\% & \bf 24.81 & \bf 40.15 & \bf 20.04\% \\\cline{2-14}
				 & MTGNN  & 37.75 & 58.37 & 35.17\% & 46.93 & 72.20 & 43.67\% & 59.01 & 88.85 & 54.35\% & 46.72 & 71.38 & 43.07\% \\
				& \bf +PCA  & \bf 21.03 & \bf 34.05 & \bf 16.85\% & \bf 24.18 & \bf 38.86 & \bf 19.35\% & \bf 30.20 & \bf 47.96 & \bf 24.54\% & \bf 24.61 & \bf 39.45 & \bf 19.74\% \\\cline{2-14}
				& GWNET  & 23.53 & 37.60 & 21.39\% & 28.30 & 44.73 & 25.43\% & 36.33 & 56.29 & 31.47\% & 28.63 & 45.07 & 25.29\% \\
				& \bf +PCA  & \bf 21.46 & \bf 34.81 & \bf 20.68\% & \bf 25.10 & \bf 40.29 & \bf 24.18\% & \bf 31.70 & \bf 49.59 & \bf 29.95\% & \bf 25.51 & \bf 40.69 & \bf 24.29\% \\\cline{2-14}
				& DGCRN  & 32.71 & 49.62 & 30.02\% & 39.38 & 60.64 & 36.73\% & 49.94 & 76.55 & 45.68\% & 39.58 & 60.58 & 36.49\% \\
				& \bf +PCA  & \bf 21.83 & \bf 34.92 & \bf 18.40\% & \bf 25.39 & \bf 40.33 & \bf 21.94\% & \bf 31.77 & \bf 49.54 & \bf 27.96\% & \bf 25.75 & \bf 40.72 & \bf 22.13\% \\\cline{2-14}
				& STAEFORMER  & 27.80 & 43.06 & 24.38\% & 33.94 & 51.94 & 29.21\% & 43.74 & 65.82 & 37.73\% & 34.25 & 52.30 & 29.66\% \\
				& \bf +PCA  & \bf 22.05 & \bf 35.82 & \bf 19.33\% & \bf 25.76 & \bf 41.61 & \bf 22.38\% & \bf 32.53 & \bf 52.13 & \bf 27.41\% & \bf 26.18 & \bf 42.25 & \bf 22.47\% \\\cline{2-14}
				& D$^2$STGNN  & 28.22 & 44.75 & 25.42\% & 34.96 & 55.48 & 30.63\% & 45.13 & 70.44 & 38.30\% & 35.23 & 55.60 & 30.64\% \\
				& \bf +PCA  & \bf 21.18 & \bf 34.38 & \bf 17.56\% & \bf 24.58 & \bf 39.46 & \bf 21.64\% & \bf 30.06 & \bf 47.64 & \bf 27.45\% & \bf 24.74 & \bf 39.68 & \bf 21.63\% \\ \bottomrule
				 & \cellcolor{green!30} Improve 
				  & \cellcolor{green!30}$\textbf{26.3\%}\downarrow$ 
				  & \cellcolor{green!30}$\textbf{29.6\%}\downarrow$ 
				  & \cellcolor{green!30}$\textbf{30.1\%}\downarrow$ 
				  & \cellcolor{green!30}$\textbf{24.1\%}\downarrow$ 
				  & \cellcolor{green!30}$\textbf{28.0\%}\downarrow$ 
				  & \cellcolor{green!30}$\textbf{28.8\%}\downarrow$ 
				  & \cellcolor{green!30}$\textbf{27.6\%}\downarrow$ 
				  & \cellcolor{green!30}$\textbf{28.8\%}\downarrow$ 
				  & \cellcolor{green!30}$\textbf{27.4\%}\downarrow$ 
				  & \cellcolor{green!30}$\textbf{28.7\%}\downarrow$ 
				  & \cellcolor{green!30}$\textbf{27.0\%}\downarrow$ 
				  & \cellcolor{green!30}$\textbf{27.9\%}\downarrow$ \\

				\bottomrule
			\end{tabular}
		}
	\end{sc}
\end{table*}

\begin{table*}[t]
	\centering
	\small
	\renewcommand{\arraystretch}{1.3}
	\tabcolsep=1mm
	\caption{Performance comparisons are presented, with the best-performing baseline results highlighted in bold.  "Param" denotes the number of learnable parameters, where K represents thousands ($10^3$) and M represents millions ($10^6$).}\label{tab:performance2}
	\begin{sc}
		\resizebox{\textwidth}{!}{
			\begin{tabular}{lcccc|ccc|ccc|ccc}
				\toprule
				\multirow{2}{*}{Data} & \multirow{2}{*}{Method} & \multicolumn{3}{c}{Horizon 3} & \multicolumn{3}{c}{Horizon 6} & \multicolumn{3}{c}{Horizon 12} & \multicolumn{3}{c}{Average} \\ \cline{3-14} 
				&    & MAE & RMSE & MAPE & MAE & RMSE & MAPE & MAE & RMSE & MAPE & MAE & RMSE & MAPE \\ 
				\hline \hline
				\multirowcell{13}{\bf PEMS07 \\ \bf 2017 \\ $\downarrow$ \\ \bf 2018} & AGCRN & 36.53 & 50.06 & 52.61\% & 50.46 & 67.40 & 79.76\% & 77.05 & 100.05 & 144.69\% & 52.36 & 69.74 & 86.04\%    \\
				& \bf +PCA & \bf 33.17 & \bf 42.14 & \bf 46.33\% & \bf 45.57 & \bf 62.92 & \bf 60.04\% & \bf 57.18 & \bf 82.44 & \bf 98.94\% & \bf 40.75 & \bf 61.76 & \bf 77.99\% \\ \cline{2-14}
				& TrendGCN & 27.05 & 39.39 & 19.35\% & 35.75 & 50.57 & 37.64\% & 54.98 & 76.28 & 77.52\% & 37.70 & 53.33 & 41.85\%   \\
				& \bf +PCA & \bf 20.44 & \bf 31.43 & \bf 14.92\%  & \bf 21.85 & \bf 33.73 & \bf 16.65\%  & \bf 23.56 & \bf 36.40 & \bf 16.75\%  & \bf 21.66 & \bf 33.45 & \bf 15.26\% \\ \cline{2-14}
				& STID & 29.60 & 43.07 & 21.88\% & 38.09 & 54.60 & 37.68\% & 53.36 & 74.79 & 71.10\% & 38.96 & 55.61 & 41.11\%          \\
				& \bf +PCA & \bf 21.62 & \bf 33.09 & \bf 11.35\%  & \bf 24.13 & \bf 36.83 & \bf 12.90\%  & \bf 28.51 & \bf 42.79 & \bf 16.53\%  & \bf 24.29 & \bf 36.88 & \bf 13.40\% \\ \cline{2-14}
				& GWNeT & 38.43 & 53.72 & 53.63\% & 52.04 & 73.10 & 68.82\% & 72.04 & 100.98 & 107.42\% & 52.36 & 73.42 & 73.31\%\\
				& \bf +PCA & \bf 22.43 & \bf 34.11 & \bf 16.05\%  & \bf 25.69 & \bf 38.80 & \bf 17.50\%  & \bf 31.11 & \bf 46.15 & \bf 21.31\%  & \bf 25.82 & \bf 38.86 & \bf 17.75\% \\ \cline{2-14}
				& MTGNN & 44.99 & 63.67 & 90.30\% & 56.54 & 79.17 & 102.14\% & 71.68 & 99.35 & 142.94\% & 56.17 & 78.82 & 108.25\%    \\
				& \bf +PCA & \bf 28.96 & \bf 41.82 & \bf 20.84\% & \bf 37.95 & \bf 54.64 & \bf 31.96\% & \bf 48.58 & \bf 68.03 & \bf 55.61\% & \bf 37.91 & \bf 54.28 & \bf 33.78\% \\ \cline{2-14}
				& DGCRN & 28.81 & 41.09 & 31.88\% & 36.33 & 52.00 & 42.87\% & 48.62 & 69.43 & 56.72\% & 36.91 & 52.77 & 43.50\%           \\
				& \bf +PCA & \bf 28.14 & \bf 40.76 & \bf 22.79\% & \bf 30.79 & \bf 44.39 & \bf 24.19\% & \bf 33.83 & \bf 48.88 & \bf 26.25\% & \bf 30.93 & \bf 44.54 & \bf 24.09\% \\ \cline{2-14}
				& STAEformer & 31.30 & 45.85 & 28.27\% & 41.53 & 60.07 & 34.54\% & 59.68 & 84.25 & 56.29\% & 42.92 & 61.68 & 37.81\%       \\
				& \bf +PCA & \bf 19.90 & \bf 30.87 & \bf 14.46\%  & \bf 21.34 & \bf 33.10 & \bf 16.46\%  & \bf 23.37 & \bf 35.82 & \bf 20.60\%  & \bf 21.27 & \bf 32.84 & \bf 17.08\% \\ \cline{2-14}
				& D$^2$STGNN & 27.57 & 40.87 & 15.09\% & 35.74 & 53.42 & 18.84\% & 49.74 & 75.27 & 26.21\% & 36.51 & 54.79 & 19.74\%   \\
				& \bf +PCA & \bf 20.85 & \bf 32.05 & \bf 9.88\%  & \bf 22.82 & \bf 35.05 & \bf 11.66\%  & \bf 26.81 & \bf 40.54 & \bf 13.25\%  & \bf 23.19 & \bf 35.40 & \bf 11.98\% \\  \bottomrule
				& \cellcolor{green!30} Improve 
				& \cellcolor{green!30}$\textbf{9.2\%}\downarrow$ 
				& \cellcolor{green!30}$\textbf{9.7\%}\downarrow$ 
				& \cellcolor{green!30}$\textbf{25.8\%}\downarrow$ 
				& \cellcolor{green!30}$\textbf{15.8\%}\downarrow$ 
				& \cellcolor{green!30}$\textbf{6.6\%}\downarrow$ 
				& \cellcolor{green!30}$\textbf{17.6\%}\downarrow$ 
				& \cellcolor{green!30}$\textbf{11.9\%}\downarrow$ 
				& \cellcolor{green!30}$\textbf{24.7\%}\downarrow$ 
				& \cellcolor{green!30}$\textbf{31.6\%}\downarrow$ 
				& \cellcolor{green!30}$\textbf{14.9\%}\downarrow$ 
				& \cellcolor{green!30}$\textbf{13.3\%}\downarrow$ 
				& \cellcolor{green!30}$\textbf{22.7\%}\downarrow$ \\
				
				\midrule\midrule
				\multirowcell{13}{\bf PEMS08 \\ \bf 2016 \\ $\downarrow$ \\ \bf 2017} & AGCRN  & 38.88 & 53.73 & 44.53\% & 48.51 & 67.49 & 55.89\% & 61.35 & 84.68 & 69.50\% & 48.87 & 67.71 & 56.15\% \\
				& \bf +PCA  & \bf 27.75 & \bf 42.22 & \bf 28.60\% & \bf 32.22 & \bf 47.63 & \bf 34.71\% & \bf 40.18 & \bf 57.62 & \bf 42.07\% & \bf 32.77 & \bf 48.41 & \bf 34.28\% \\\cline{2-14}
				& TrendGCN  & 27.57 & 39.98 & 27.29\% & 35.83 & 52.52 & 35.73\% & 48.22 & 69.42 & 47.86\% & 36.05 & 52.41 & 35.77\% \\
				& \bf +PCA  & \bf 19.62 & \bf 29.59 & \bf 18.37\% & \bf 24.40 & \bf 37.32 & \bf 24.20\% & \bf 31.76 & \bf 47.79 & \bf 34.51\% & \bf 24.61 & \bf 37.28 & \bf 24.95\% \\\cline{2-14}
				& STID  & 26.35 & 42.95 & 17.07\% & 36.65 & 61.68 & 24.73\% & 53.70 & 89.24 & 41.21\% & 37.33 & 62.12 & 26.36\% \\
				& \bf +PCA  & \bf 14.14 & \bf 23.32 & \bf 8.39\% & \bf 15.82 & \bf 26.58 & \bf 9.43\% & \bf 18.92 & \bf 31.65 & \bf 11.73\% & \bf 15.96 & \bf 26.62 & \bf 9.60\% \\\cline{2-14}
				 & MTGNN  & 38.29 & 52.91 & 61.07\% & 50.77 & 70.24 & 84.82\% & 69.05 & 95.04 & 115.27\% & 50.95 & 70.36 & 82.79\% \\
				& \bf +PCA  & \bf 21.66 & \bf 32.93 & \bf 11.06\% & \bf 24.43 & \bf 37.08 & \bf 12.69\% & \bf 29.51 & \bf 44.04 & \bf 17.54\% & \bf 24.66 & \bf 37.24 & \bf 13.36\% \\\cline{2-14}
				& GWNET  & 20.09 & 30.16 & 16.83\% & 25.41 & 38.14 & 20.32\% & 33.50 & 49.41 & 25.78\% & 25.56 & 38.16 & 20.19\% \\
				& \bf +PCA  & \bf 15.70 & \bf 25.25 & \bf 9.16\% & \bf 17.94 & \bf 29.16 & \bf 10.48\% & \bf 22.14 & \bf 35.49 & \bf 13.81\% & \bf 18.14 & \bf 29.25 & \bf 10.82\% \\\cline{2-14}
				& DGCRN  & 29.82 & 45.15 & 32.24\% & 38.22 & 59.26 & 42.76\% & 49.21 & 75.64 & 55.96\% & 37.81 & 58.08 & 42.14\% \\
				& \bf +PCA  & \bf 16.06 & \bf 25.36 & \bf 11.47\% & \bf 18.74 & \bf 29.75 & \bf 13.84\% & \bf 23.41 & \bf 36.75 & \bf 18.31\% & \bf 18.93 & \bf 29.85 & \bf 14.14\% \\\cline{2-14}
				& STAEFORMER  & 24.40 & 36.42 & 19.10\% & 30.16 & 44.91 & 23.22\% & 39.87 & 58.84 & 31.95\% & 30.50 & 45.27 & 23.99\% \\
				& \bf +PCA  & \bf 13.59 & \bf 23.54 & \bf 8.94\% & \bf 15.19 & \bf 26.81 & \bf 9.72\% & \bf 18.14 & \bf 31.91 & \bf 11.63\% & \bf 15.34 & \bf 26.88 & \bf 9.88\% \\\cline{2-14}
				& D$^2$STGNN  & 27.43 & 41.77 & 28.42\% & 38.01 & 58.12 & 38.73\% & 51.92 & 77.25 & 51.78\% & 38.99 & 59.36 & 38.97\% \\
				& \bf +PCA  & \bf 14.74 & \bf 24.22 & \bf 8.50\% & \bf 16.92 & \bf 28.00 & \bf 9.62\% & \bf 21.25 & \bf 34.69 & \bf 12.01\% & \bf 17.26 & \bf 28.34 & \bf 9.92\% \\ \bottomrule
				& \cellcolor{green!30} Improve 
				& \cellcolor{green!30}$\textbf{28.6\%}\downarrow$ 
				& \cellcolor{green!30}$\textbf{33.6\%}\downarrow$ 
				& \cellcolor{green!30}$\textbf{34.5\%}\downarrow$ 
				& \cellcolor{green!30}$\textbf{21.4\%}\downarrow$ 
				& \cellcolor{green!30}$\textbf{29.4\%}\downarrow$ 
				& \cellcolor{green!30}$\textbf{32.0\%}\downarrow$ 
				& \cellcolor{green!30}$\textbf{35.8\%}\downarrow$ 
				& \cellcolor{green!30}$\textbf{37.9\%}\downarrow$ 
				& \cellcolor{green!30}$\textbf{39.5\%}\downarrow$ 
				& \cellcolor{green!30}$\textbf{32.2\%}\downarrow$ 
				& \cellcolor{green!30}$\textbf{27.6\%}\downarrow$ 
				& \cellcolor{green!30}$\textbf{37.7\%}\downarrow$ \\
				\bottomrule
			\end{tabular}
		}
	\end{sc}
\end{table*}

\begin{figure*}[t]
	\centering
	\includegraphics[width=0.85\linewidth]{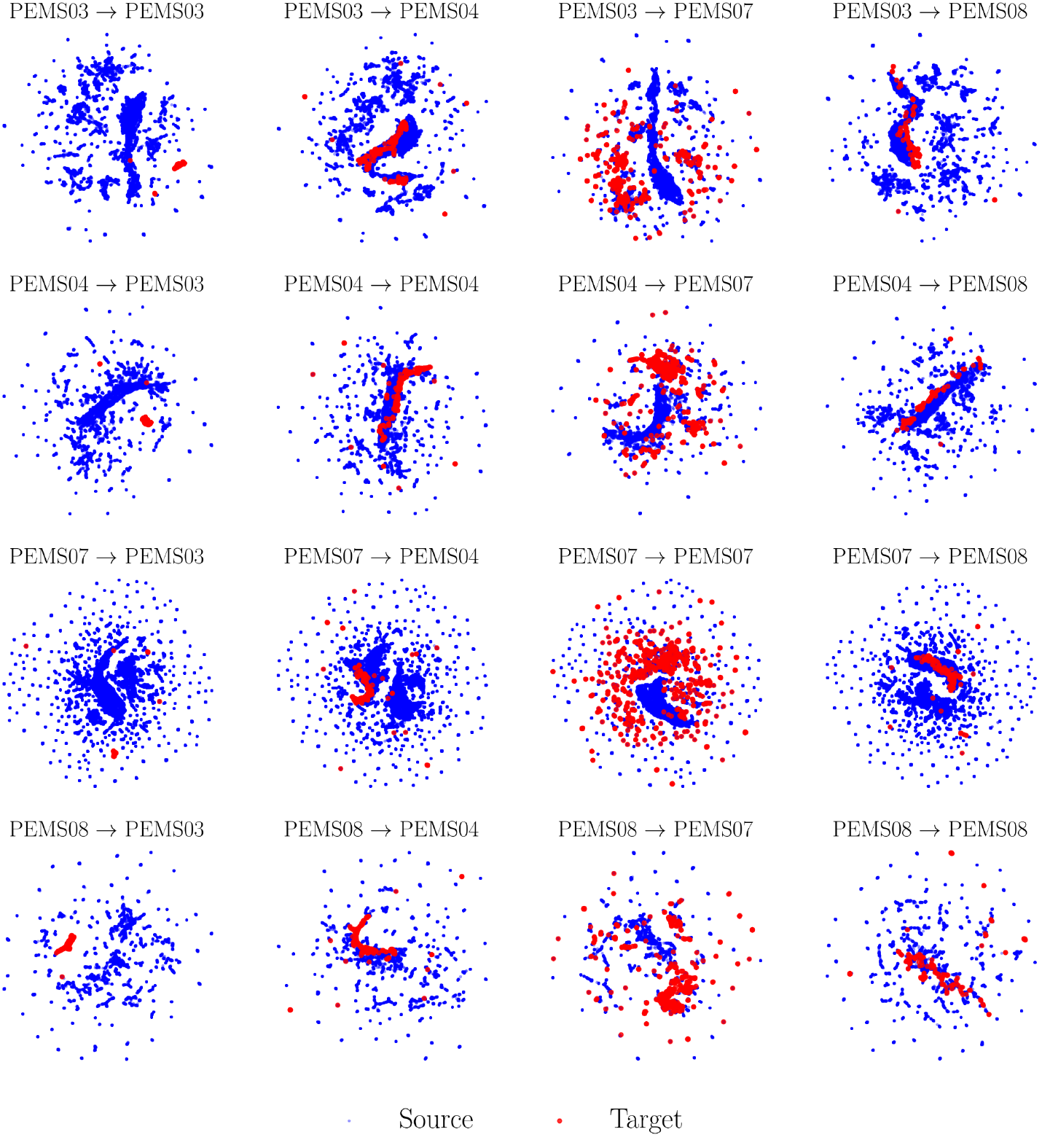}
	\caption{\textbf{Visualization of pairwise PCA-embedding alignment for traffic data}: Each scatter plot illustrates traffic pattern comparisons across different PEMS datasets, where blue points represent source dataset embeddings and red points represent target dataset embeddings. The 4×4 matrix demonstrates cross-dataset embedding relationships, with diagonal plots showing within-dataset distributions and off-diagonal plots revealing cross-dataset transferability patterns. The distribution overlaps between source and target embeddings indicate the potential for zero-shot generalization and structural similarities among different traffic monitoring systems. This visualization validates how PCA embeddings can effectively capture and align shared traffic patterns across diverse geographical contexts, facilitating cross-dataset knowledge transfer in traffic prediction tasks.}
	\label{fig:pcadim}
\end{figure*}
\subsection{Comparsion with  Adaptive Embedding (RQ1)} 
The primary research question we aimed to address was whether the use of PCA embeddings, as opposed to learnable embeddings, would lead to a decline in model performance during training. To explore this, we conducted a comparative analysis of model performance on the PEMS benchmark, utilizing both PCA embeddings and learnable embeddings. 

As shown in \figref{fig:fix_perf}, our results demonstrate that the use of PCA embeddings does not significantly reduce model effectiveness compared to learnable embeddings, suggesting that manual dimensionality reduction can achieve outcomes comparable to those produced by learnable approaches. Notably, in certain cases, PCA embeddings even exhibited superior performance. We hypothesize that this improvement is largely due to the greater ability of PCA embeddings to mitigate overfitting in the model. Recent studies \cite{CaST,zhou2023maintaining,ji2023self} have identified spatial-shift issues, despite the original test data being collected only three weeks apart. We observe that recurrent neural network architectures, such as AGCRN and TrendGCN, are more prone to overfitting, and transformer architectures also encounter this issue on the PEMS03 dataset.

\subsection{Zero-Shot Performance (RQ2)} While the utilization of PCA embeddings does not necessarily yield improvements in in-distribution performance, our work has unveiled a significant advantage in terms of model interpretability. It allows us to transcend previous training paradigms, which typically confined model validation to identical test sets. Our approach enables model validation across diverse datasets, irrespective of variations in node numbers. 
Specifically, as illustrated in the \tableref{tab:trans}, we employed PCA embeddings for training on the PEMS03 and PEMS07 datasets.  We use $A \rightarrow B$ to indicate that the model is trained on the dataset $A$  and tested on the dataset $B$. During the testing phase, we substituted the embeddings by applying the projection matrices $W$ derived from PEMS03 and PEMS07, respectively, to generate the embedding $E$ of the training samples from PEMS04 and PEMS08. 

\tableref{tab:trans} reveal that STAEformer (transformer-based architectures) exhibit remarkable zero-shot generalization capabilities.
Furthermore, our findings indicate that leveraging larger datasets, such as PEMS07's 883 nodes compared to PEMS03's 358, leads to superior performance, which suggests a positive correlation between dataset size and generalization capacity.
Based on these results, we posit that PCA embeddings may emerge as a unifying paradigm for future large-scale traffic models.

Similarly, we conducted experiment in LargeST \cite{liu2024largest} in \tableref{tab:trans}. In a manner consistent with the previous experiments, we trained the model separately on the San Diego and Bay Area datasets, then tested it on the remaining scenarios. We observed that the zero-shot capabilities of the ST-Model were significantly underestimated. The previous training paradigms did not fully unlock the model’s potential. In contrast, our results demonstrate that the zero-shot performance is not substantially inferior to that of the model trained on these datasets. We believe that the PCA-embedding approach paves the way for the development of future large-scale models for transportation.

\subsection{Test Performance on PEMS over a Year (RQ3)}
Tables \ref{tab:performance1} and \ref{tab:performance2} present a comparative analysis of the performance of ST-models trained on the original PEMS benchmark dataset \cite{guo2019attention} and tested on our collected cross-year data. We also showcase results incorporating our PCA embeddings strategy across different forecasting horizons (Horizon 3, Horizon 6, Horizon 12).
In nearly all cases, the addition of PCA embeddings significantly enhances model generalization. Our method adapts well to models of varying architectures and demonstrates excellent performance improvements across all datasets.
The improvement is particularly notable on the PEMS03 dataset, where we observe substantial performance gains. For instance, with STID, the average MAE decreased from 33.99 to 9.18, RMSE from 46.70 to 13.75, and MAPE from 686.25
On the PEMS04 dataset, we observe consistent improvements across all models. For example, with MTGNN, the average MAE decreased from 46.72 to 24.61, RMSE from 71.38 to 39.45, and MAPE from 43.07
For PEMS07 and PEMS08, the improvements are equally impressive. In PEMS08, STID with PCA achieved remarkable gains, with average MAE decreasing from 37.33 to 15.96, RMSE from 62.12 to 26.62, and MAPE from 26.36
These comprehensive improvements across different datasets and model architectures demonstrate the effectiveness of our PCA embeddings strategy in enhancing spatiotemporal traffic forecasting performance. The consistent reduction in MAE, RMSE, and MAPE metrics across various prediction horizons underscores the robustness and generalizability of our approach.

\subsection{Comparison with Different Strategies (RQ4)}
We extend our analysis by conducting a comprehensive comparison between the PCA-Embedding and the zero embedding strategy in \tableref{tab:res2}. The zero embedding strategy, which involves setting the adaptive embedding to zero during the testing phase, aims to eliminate potential biases introduced during the training phase \cite{STID}. However, our empirical observations suggest that this approach is suboptimal for zero-shot evaluation when compared to using the original embeddings. 
Moreover, we examined a fine-tuning strategy for the adaptive embedding, where fine-tuning was conducted using the same 5\% of validation data employed for the PCA. The comparative analysis highlights substantial differences in generalization performance across various architectures. Notably, simpler embedding strategies, such as those utilized by STID and STAEformer, demonstrate superior generalization capabilities. The PCA embedding, in particular, yields performance metrics that are closely aligned with those achieved through fine-tuning. 
This outcome can be rationalized by considering the inherent purpose of adaptive embedding, as articulated in STID, which is primarily to mitigate the problem of spatial-temporal indistinguishability. By applying PCA, the principal component effectively delineates representations of different sensors, thereby enhancing the overall discriminative power of the model. The result suggests that PCA might be sufficient for a broad range of scenarios, especially when computational efficiency and model interpretability are prioritized.

\subsection{Visualization of PCA-embedding (RQ4)}

\figref{fig:pcadim} presents a comprehensive 4×4 matrix of PCA-embedding alignments, illustrating the cross-dataset transferability patterns among different PEMS datasets. Each subplot demonstrates the distribution of PCA embeddings in a dimensionally-reduced feature space, where blue points represent the source dataset embeddings and red points indicate target dataset embeddings. The visualization methodology effectively illuminates the inherent structural correspondences and divergences among distinct traffic monitoring systems. The spatial arrangement of embeddings reveals varying degrees of cross-dataset alignment, which is particularly crucial for understanding the potential for transfer learning and zero-shot generalization in traffic prediction tasks. Notably, certain dataset pairs exhibit high degrees of embedding overlap, suggesting robust structural similarities in their underlying traffic patterns, while others display more disparate distributions, indicating domain-specific characteristics. This PCA-based embedding analysis not only validates the presence of transferable features across different PEMS datasets but also provides a quantitative framework for assessing the feasibility of cross-domain adaptation in intelligent transportation systems, thereby contributing to our understanding of the fundamental structures underlying urban traffic patterns and their potential for generalization across different monitoring systems.

\begin{figure}[t]
	\centering
	\subfigure[STID]{\includegraphics[width=0.48\linewidth]{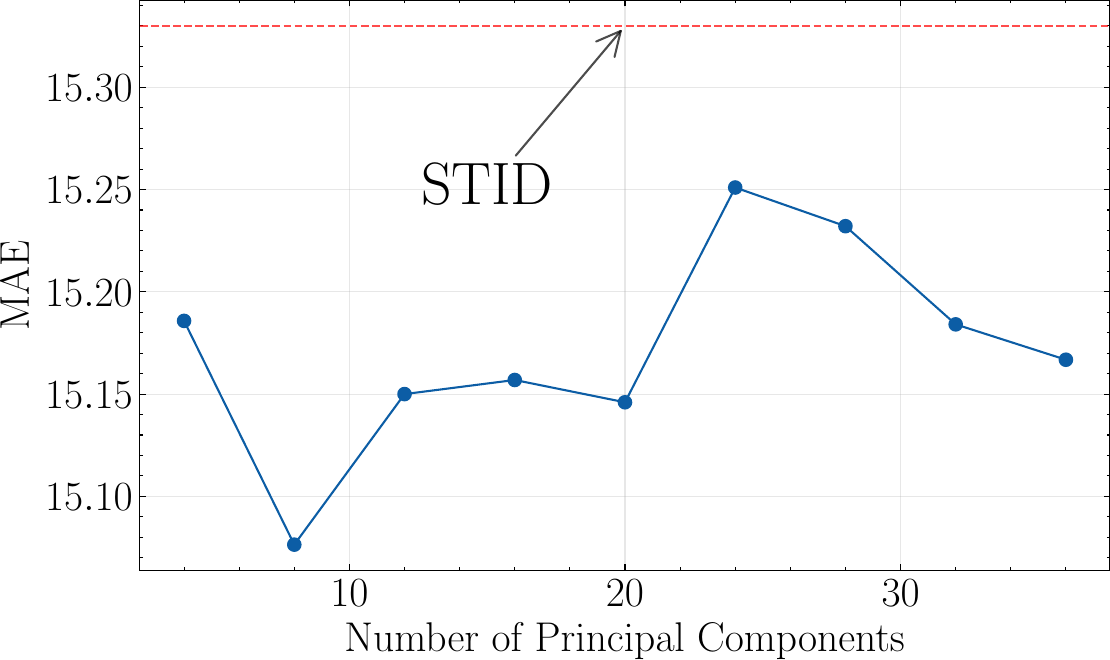}}
	\subfigure[STAEformer]{\includegraphics[width=0.48\linewidth]{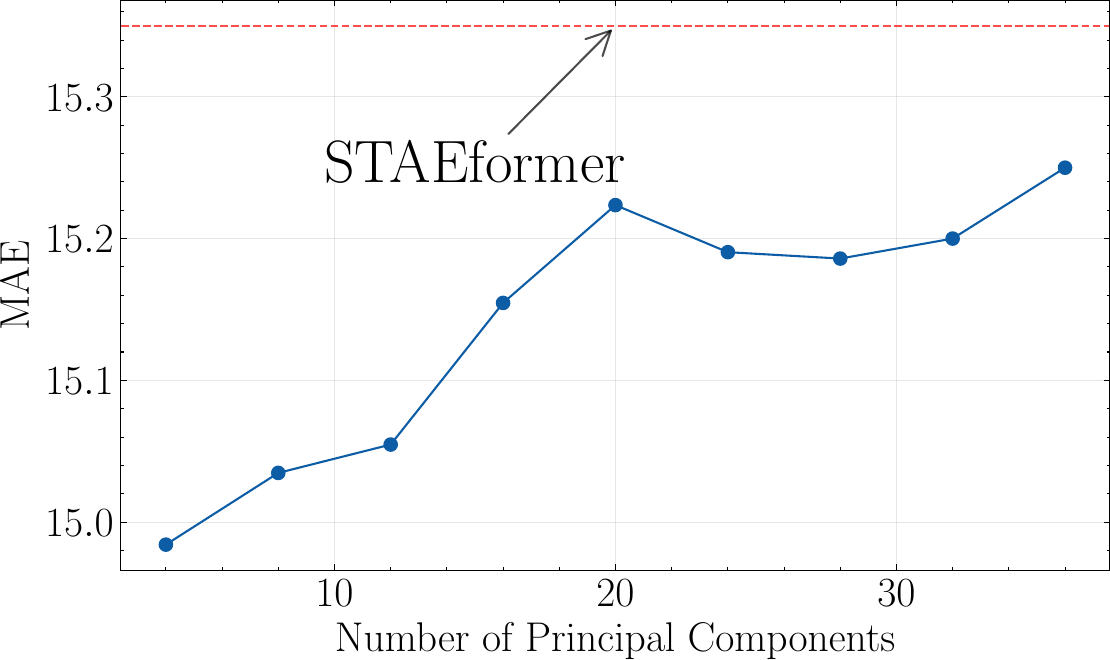}}
	\caption{\textbf{Component search results for the optimal number of principal components in STID and STAEformer when applied to the PEMS03 dataset are presented.} The red dashed lines signify the baseline performance. The fluctuating MAE curves indicate that traditional methods are susceptible to overfitting. However, by employing PCA embeddings and utilizing only a subset of principal components, it is possible to retain most of the information while achieving superior performance.}
	\label{fig:ablationsearchstidpems03}
\end{figure}

\subsection{Can PCA Embedding avoid over-distinguishability? (RQ5)}
We conducted a comprehensive grid search to determine the optimal number of principal components for the STID and STAEformer models when applied to the PEMS03 dataset, as visualized in Figure \ref{fig:ablationsearchstidpems03}. Our results suggest that both models achieve peak performance with approximately 4 and 8 principal components, respectively. Beyond these thresholds, performance either deteriorates (in the case of STID) or plateaus (for STAEformer). The fluctuating Mean Absolute Error (MAE) curves, particularly evident in STID, indicate that traditional methods may be susceptible to overfitting when incorporating an excessive degree of spatial differentiation. Our findings demonstrate that using PCA embeddings with a constrained number of principal components can effectively preserve crucial spatial information while surpassing the baseline model that relies on trainable adaptive embeddings (represented by the red dashed lines). This outcome corroborates previous mention that has cautioned against the potential pitfalls of excessive distinguishability in adaptive embeddings, especially in datasets where spatial relationships are paramount.

\section{Related Work}
\subsection{Adaptive Embeddings in Spatiotemporal Models.} Adaptive Embeddings address the limitations of traditional graph construction by using adaptive learning strategies to uncover implicit relationships within spatiotemporal data. Existing approaches can be divided into random initialization-based and feature initialization-based methods.
In the random initialization approach, GWNet~\cite{GWNet} introduced adaptive adjacency matrices via two learnable embedding matrices, laying the foundation for later work. MT-GNN~\cite{MTGNN} expanded this idea by incorporating nonlinear transformations and antisymmetric operations to capture more complex relationships. CCRNN~\cite{CCRNN} and DMSTGCN~\cite{DMSTGCN} introduced layer-wise adaptive graph learning and tensor decomposition-based methods, respectively, enhancing the model's ability to capture dynamic spatial dependencies. More recently, STID~\cite{STID} and STAEformer~\cite{STAEformer} simplified the process by directly optimizing Adaptive Embeddings, bypassing graph construction while maintaining competitive performance.
Feature initialization methods have shown promising results in capturing dynamic spatial relationships. DGCRN~\cite{li2021dynamic} employs a recurrent mechanism to dynamically generate graph structures based on hidden states, enabling the model to adapt to temporal variations. GTS~\cite{GTS} advances this approach by utilizing a probabilistic framework to generate graph structures from input features, enabling more data-driven and adaptive modeling. These methods demonstrate superior performance in scenarios where spatial relationships evolve significantly over time.
\subsection{Spatiotemporal Forecasting.}
Spatiotemporal forecasting has been extensively researched due to its crucial role in various real-world applications~\cite{liu2024largest,MemeSTN,SpatiotemporalDiffusionPointProcesses,liuhao2023,Greto,CDSTG,GMRL,MoSSL}. Traditional methods, including ARIMA~\cite{ARIMA-traffic}, VAR~\cite{VAR}, $k$-NN~\cite{kNN-traffic}, and SVM~\cite{SVM-traffic}, often fail to capture the complex spatiotemporal dependencies in data due to their inherent limitations in modeling non-linear relationships and high-dimensional feature spaces.
In recent years, GCNs have been integrated with temporal models, leading to improved performance in spatiotemporal forecasting. Notable examples include STGCN~\cite{STGCN}, which combines graph convolutions with gated temporal convolutions, and DCRNN~\cite{DCRNN}, which integrates diffusion convolution with recurrent neural networks. These pioneering works have inspired numerous developments in the field~\cite{GWNet,AGCRN,StemGNN,STID}. Additionally, innovative methods have been proposed to address specific challenges in spatiotemporal forecasting~\cite{GTS,DSTAGNN,PMMemNet,MegaCRN,CaST,jin2023survey}, such as long-term dependencies, multi-scale temporal patterns, and heterogeneous spatial relationships. However, current research primarily evaluates ST-GNNs within short timeframes, neglecting the dynamics of data distribution shifts and long-term pattern changes.
\subsection{Distribution Shift in GNNs.} Graph Neural Networks have significantly advanced graph representation learning, achieving state-of-the-art results across various graph-related tasks~\cite{kipf2016semi,gat,hamilton2017inductive,chen2021gpr,klicpera2018predict-appnp,wu2019comprehensive-survey}. However, recent studies have revealed suboptimal performance of GNNs on out-of-distribution data for both node and graph classification tasks~\cite{zhu2021shift,wu2022handling,liu2022confidence,chen2022invariance,buffelli2022sizeshiftreg,gui2022good,wu2022discovering,you2023graph}. These findings highlight the vulnerability of GNNs to distribution shifts and the importance of developing robust models.
Recent pioneering efforts~\cite{CaST,zhou2023maintaining,ji2023self} have made significant strides in addressing the challenge of out-of-distribution scenarios in traffic forecasting. CaST~\cite{CaST} introduces a causal structure learning framework to enhance model robustness, while~\cite{zhou2023maintaining} proposes invariant learning techniques to maintain prediction stability. However, these studies either manually construct the spatial-shift or validate it only within short-term scenarios, limiting their applicability to real-world situations. Although distribution shift has been observed across various fields, in the domain of traffic prediction, spatial-shift has yet to be widely validated in real-world scenarios, particularly in the context of long-term forecasting and evolving urban environments.
The gap between theoretical advances in handling distribution shift and practical applications in traffic forecasting remains significant, particularly in scenarios involving long-term temporal evolution and complex spatial dynamics. This highlights the need for more comprehensive approaches that can effectively address both spatial and temporal distribution shifts in real-world traffic forecasting applications.

\section{Conclusion}
We reveal the limitations of current ST-models in adapting to evolving urban spatial relationships, leading to spatial-shift performance degradation in traffic forecasting. We propose a novel, PCA embedding enabling training-free adaptation to new scenarios. Our approach, implemented in existing architectures, significantly improves prediction accuracy and generalization capability. It also demonstrates potential for cross-dataset zero-shot predictions.
Our work advances traffic forecasting methodology and addresses challenges posed by rapid urbanization. 
Future work will focus on refining the adaptive framework, exploring its applicability to other dynamic prediction tasks.

\bibliographystyle{IEEEtran}
\bibliography{reference}
\begin{IEEEbiography}[{\includegraphics[width=1in,height=1.25in,clip,keepaspectratio]{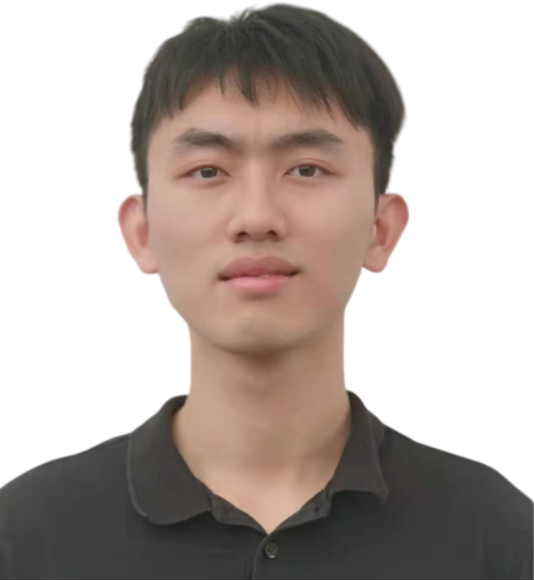}}]{Hongjun Wang} is working toward the PhD degree in the Department of Mechano-Informatics at The University of Tokyo. He received his M.S. degree in computer science and technology from Southern University of Science and Technology, China. He received his B.E. degree from the Nanjing University of Posts and Telecommunications, China, in 2019. His research interests are broadly in machine learning, with a focus on urban computing, explainable AI, data mining, and data visualization.
\end{IEEEbiography}
\vspace{1ex}
\begin{IEEEbiography}[{\includegraphics[width=1in,height=1.25in,clip,keepaspectratio]{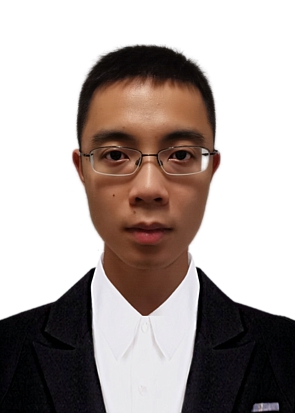}}]{Jiyuan Chen} is working towards his PhD degree at The Hong Kong Polytechnic University. He received his B.S. degree in Computer Science and Technology from Southern University of Science and Technology, China. His major research fields include artificial intelligence, deep learning, urban computing, and data mining.
\end{IEEEbiography}

\vspace{1ex}
\begin{IEEEbiography}[{\includegraphics[width=1in,height=1.25in,clip,keepaspectratio]{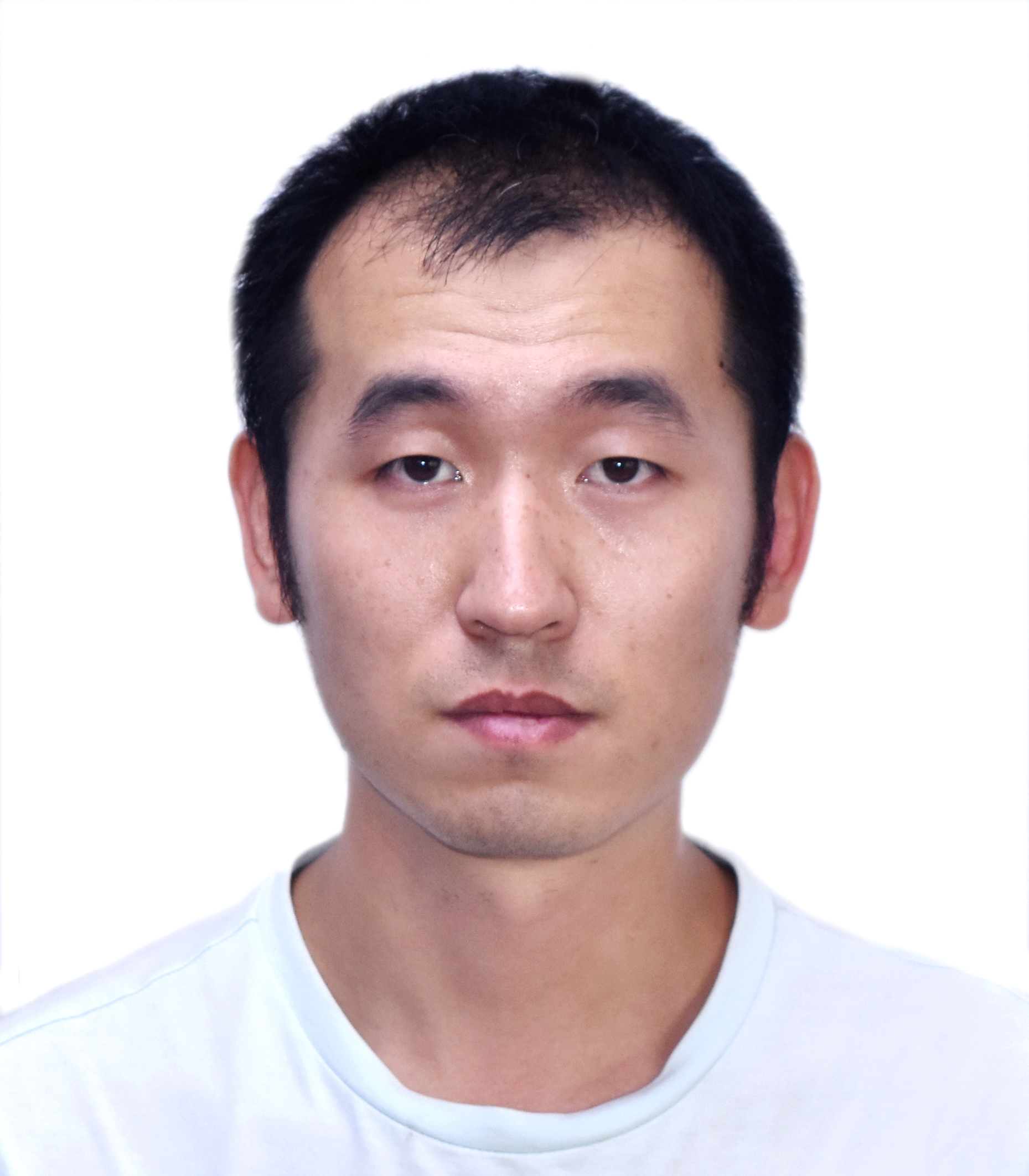}}]{Lingyu Zhang}   joined Baidu in 2012 as a search strategy algorithm research and development engineer. He joined Didi in 2013 and served as senior algorithm engineer, technical director of taxi strategy algorithm direction, and technical expert of strategy model department. Currently a researcher at Didi AI Labs, he used machine learning and big data technology to design and lead the implementation of multiple company-level intelligent system engines during his work at Didi, such as the order distribution system based on combination optimization, and the capacity based on density clustering and global optimization. Scheduling engine, traffic guidance and personalized recommendation engine, "Guess where you are going" personalized destination recommendation system, etc. Participated in the company's dozens of international and domestic core technology innovation patent research and development, application, good at using mathematical modeling, business model abstraction, machine learning, etc. to solve practical business problems. He has won honorary titles such as Beijing Invention and Innovation Patent Gold Award and QCon Star Lecturer, and his research results have been included in top international conferences related to artificial intelligence and data mining such as KDD, SIGIR, AAAI, and CIKM.
\end{IEEEbiography}
\vspace{1ex}
\begin{IEEEbiography}[{\includegraphics[width=1in,height=1.25in,clip,keepaspectratio]{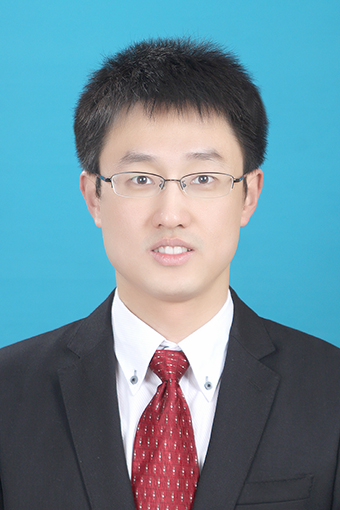}}]{Renhe Jiang} received a B.S. degree in software engineering from the Dalian University of Technology, China, in 2012, a M.S. degree in information science from Nagoya University, Japan, in 2015, and a Ph.D. degree in civil engineering from The University of Tokyo, Japan, in 2019. From 2019, he has been an Assistant Professor at the Information Technology Center, The University of Tokyo. His research interests include ubiquitous computing, deep learning, and spatio-temporal data analysis.
\end{IEEEbiography}
\vspace{1ex}
\begin{IEEEbiography}[{\includegraphics[width=1in,height=1.25in,clip,keepaspectratio]{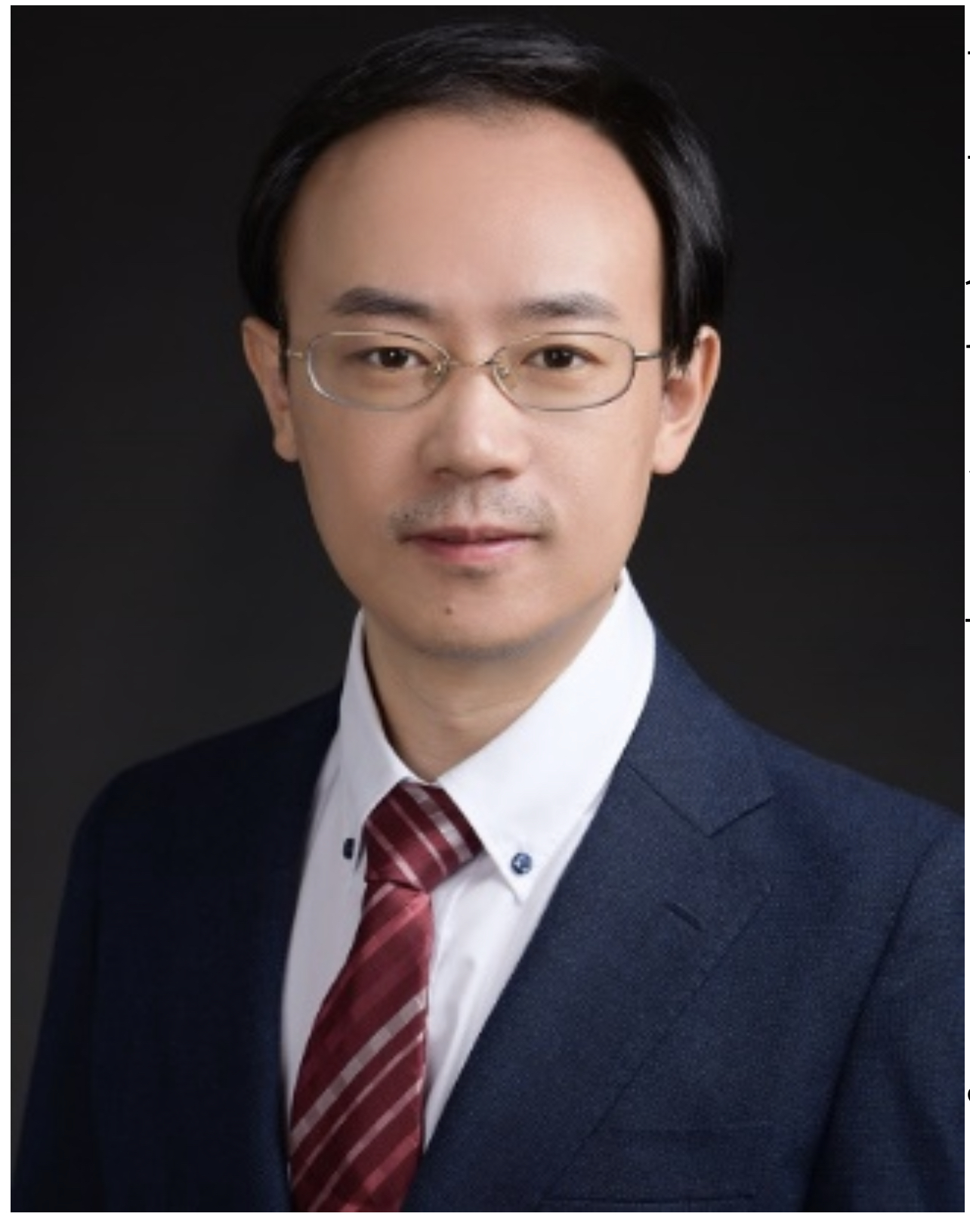}}]{Prof. Xuan Song}  received the Ph.D. degree in signal and information processing from Peking University in 2010. In 2017, he was selected as an Excellent Young Researcher of Japan MEXT. In the past ten years, he led and participated in many important projects as a principal investigator or primary actor in Japan, such as the DIAS/GRENE Grant of MEXT, Japan; Japan/US Big Data and Disaster Project of JST, Japan; Young Scientists Grant and Scientific Research Grant of MEXT, Japan; Research Grant of MLIT, Japan; CORE Project of Microsoft; Grant of JR EAST Company and Hitachi Company, Japan. He served as Associate Editor, Guest Editor, Area Chair, Program Committee Member or reviewer for many famous journals and top-tier conferences, such as IMWUT, IEEE Transactions on Multimedia, WWW Journal, Big Data Journal, ISTC, MIPR, ACM TIST, IEEE TKDE, UbiComp, ICCV, CVPR, ICRA and etc.
\end{IEEEbiography}

\end{document}